\documentclass[12pt]{article}
\usepackage{amsmath}
\usepackage{amssymb}
\usepackage{amsthm}
\usepackage{graphicx}
\usepackage{subfigure}
\usepackage{natbib}
\usepackage[]{hyperref}
\hypersetup{
    colorlinks=true,
    citecolor=purple,
    linkcolor=blue,
    filecolor=magenta,      
    urlcolor=cyan,
}

\usepackage{url} 
\usepackage{bm}
\usepackage{algorithm}
\usepackage{algpseudocode}
\usepackage{booktabs}
\usepackage{multirow}
\usepackage{diagbox}
\usepackage{hhline}
\usepackage{comment}
\usepackage{xcolor}

\newcommand{\blind}{0}

\newtheorem{theorem}{Theorem}

\newtheorem{lemma}{Lemma}

\newtheorem{model}{Model}
\newtheorem{definition}{Definition}

\graphicspath{{./figures/}{./}}		

\begin{document}

\bibliographystyle{apalike}

\def\spacingset#1{\renewcommand{\baselinestretch}%
{#1}\small\normalsize} \spacingset{1}


\if0\blind
{
  \title{\bf Simultaneous Dimensionality and Complexity Model Selection for Spectral Graph Clustering}
  \author{Congyuan Yang \\
    Department of Electrical and Computer Engineering, \\Johns Hopkins University\\
    \and \\
    Carey E. Priebe \\
    Department of Applied Math and Statistics, \\Johns Hopkins University
    \and \\
    Youngser Park\\
    Center for Imaging Science,\\
Johns Hopkins University \\
    \and \\
    David J. Marchette \\
    Applied Mathematics and Data Analytics Group,\\
    Naval Surface Warfare Center Dahlgren Division}
  \maketitle
} \fi

\if1\blind
{
  \bigskip
  \bigskip
  \bigskip
  \begin{center}
    {\LARGE\bf Title}
\end{center}
  \medskip
} \fi

\bigskip
\begin{abstract}
Our problem of interest is to cluster vertices of a graph by identifying underlying community structure. Among various vertex clustering approaches, spectral clustering is one of the most 
popular methods because it is easy to implement while often outperforming more
traditional clustering algorithms. However, there are two inherent model selection problems in spectral clustering, namely estimating both the embedding dimension and number of clusters. This paper attempts to address the issue by establishing a novel model selection framework specifically for vertex clustering on graphs under a stochastic block model. The first contribution is a probabilistic model which approximates the distribution of the extended spectral embedding of a graph. The model is constructed based on a theoretical result of asymptotic normality for the informative part of the embedding, and on simulation results providing a conjecture for the limiting behavior of the redundant part of the embedding. The second contribution is a simultaneous model selection framework. In contrast with traditional approaches, our model selection procedure estimates embedding dimension and number of clusters simultaneously. Based on our conjectured distributional model, a theorem on the consistency of the estimates of model parameters is presented, providing support for the utility of our method. Algorithms for our simultaneous model selection for vertex clustering are proposed, demonstrating superior performance in simulation experiments.
We illustrate our method via application to a collection of brain graphs.
\end{abstract}

\noindent%
{\it Keywords:}  Adjacency spectral embedding, stochastic block model, 
Model-based clustering.
\vfill

\newpage
\spacingset{1.5} 
\section{Introduction}
\label{sec:intro}

A mathematical graph encodes the relationships between objects in a network as edges between vertices. The analysis of such networks is of importance in many fields ranging from sociology \citep{lazer2009life} and ecology \citep{proulx2005network} to political science \citep{ward2011network} and neuroscience \citep{bullmore2011brain}. One of the most important tasks in the analysis of such a graph is to identify its community structure. This is essentially a \textit{vertex clustering} problem, in which the set of vertices is to be partitioned into nonoverlapping groups (called clusters) according to their similarities in the underlying network structure. Numerous heuristic methodologies have been proposed for vertex clustering, including divisive approaches by iteratively removing edges based on number of shortest paths \citep{girvan2002community, newman2004finding}, methods of optimizing a function called ``modularity'' which evaluates the quality of a partition \citep{bickel2009nonparametric, blondel2008fast, newman2006modularity}, algorithms employing a random walk to infer structural properties of networks \citep{pons2005computing,rosvall2008maps}, to name just a few. Among various vertex clustering approaches, we are interested here in the so-called \textit{spectral clustering} methods, for their ease of implementation, theoretical consistency and good empirical performance. 

Spectral clustering makes use of the spectral decomposition of some kind of \textit{similarity matrix} that measures the relationship between vertices. Numerous spectral clustering algorithms based on decomposing the adjacency matrix, one natural similarity matrix of the graph, have been proposed to solve the vertex clustering problem \citep{rohe2011spectral, sussman2012consistent, qin2013regularized, lei2015consistency}. Basically, the so-called \textit{adjacency spectral embedding} (ASE) is first derived by factorizing the adjacency matrix; then a traditional \textit{Gaussian mixture model} (GMM) clustering approach is applied on the ASE. Although the GMM $\circ$ ASE methods exhibit good performance, there are two inherent \textit{model selection} problems that need to be addressed in order to perform this clustering. The first is determining the number of eigenvectors whose rows are the low-dimensional points on which the GMM method is applied. Since these top eigenvectors comprise the adjacency spectral embedding, we call this number the \textit{embedding dimension}. The second is determining the number of clusters, essential for the approach of Gaussian mixture modeling.

The first model selection problem of determining the embedding dimension has received much attention over the years. In more general scenarios we call the corresponding eigenvectors "features" or "variables"; thus the problem is one of \textit{variable selection}. A comprehensive review of variable selection approaches in the model-based clustering framework has been provided in \cite{fop2018variable} and \cite{Handcock2007ModelbasedCF}. The necessity of variable selection is based on the fact that only a subset of the variables of the high-dimensional data are informative and important to the subsequent statistical inference. Using all the variables may lead to unnecessary computational cost, and may also decrease the performance of the clustering due to the irrelevance of extraneous variables. Therefore, the selection of variables which provide access to the clustering structure is of great importance. Considering the overwhelming number of methods for variable selection, we do not attempt to give a concise review of the literature. However, among various techniques for variable selection arguably the best-known methodology of principal component analysis (PCA) \citep{jolliffe2011principal} is worth mentioning. In PCA, singular values of the data matrix measure the (square root of) the variances, which reflect the importance of the variables. Consequently, the variables corresponding to $d$ top singular values are retained, as the principle components, according to the desired dimension $d$. While determining the number of principle components $d$ is an essential step before conducting PCA, we refer the readers to \cite{jackson1993stopping} for a broad review of the many so-called stopping rules. Unfortunately, there are no uniformly best rules for the task of (finite sample) dimension reduction in general due to the bias-variance tradeoff. Roughly speaking, general statistically-based heuristic approaches are often inferior in many particular applications -- universality is hard to come by, and such approaches necessarily suffer from their intended generality.
In contrast, we develop a principled approach based on 
distributional assumptions that are specific to spectral graph clustering in the stochastic block model. 

The second model selection problem, namely determining the number of clusters, is also a widely studied problem. As numerous approaches have been proposed on this topic, we refer the readers to the detailed reviews in \cite{milligan1985examination} and \cite{hardy1996number}. One substantial category of such methods is the \textit{information criterion} approach. These methods evaluate and compare a so-called information criterion, usually some kind of penalized likelihood, on finite mixture models with different number of mixture components and various model complexity specifications to perform model selection. Many information criteria have been proposed. To list a few: Akaike information criterion (AIC) \citep{akaike1998information}, Bayesian information criterion (BIC) \citep{schwarz1978estimating}, an entropy criterion (NEC) \citep{celeux1996entropy}, integrated completed likelihood (ICL) \citep{biernacki2000assessing} and cross-validated likelihood \citep{smyth2000model}. Of these, we are mostly interested in BIC, Bayesian information criterion, since it is a well-studied and easily implemented approach. Moreover, the consistency of estimation for number of components using BIC is theoretically supported in \cite{keribin2000consistent}. The practical performance of BIC-based approaches in model selection has also been highly rated by a large number of works \citep{roeder1997practical, stanford1997principal, dasgupta1998detecting, campbell1999model}.

The traditional way to address both of the model selection problems in spectral clustering is to execute the corresponding approaches sequentially. That is, one applies spectral embedding with the dimension given by some dimension reduction technique in the first step, and then one proceeds to the model selection technique on the embedded data to estimate the number of clusters in the second step. This sequential procedure of model selection suffers from two drawbacks. First, there is no one best method for estimating the embedding dimension. Even if we choose one of the modern and commonly used scree plot methods such as \citep{zhu2006automatic}, the result is not robust for limited data size. Second, the latter model selection procedure, namely estimating the number of clusters, completely depends on the result of the former one, because no information of the discarded variables will be available. This may cause an accumulation of errors when the former procedure performs poorly, even if the latter procedure is reliable. The original data is truncated before applying the clustering algorithm, which means it will not be possible to take advantage of useful information contained in the discarded dimensions to improve the clustering result. Therefore, jointly addressing these two model selection problems is desirable.

In this paper, we propose a novel simultaneous dimensionality and complexity model selection framework for spectral graph clustering. This is inspired by breakthrough work on model selection in the framework of model-based clustering proposed in \cite{raftery2006variable}. In that work, all of the variables are taken into consideration in a family of finite mixture models, which describes the distributional behavior of the raw data. The model selection procedure is conducted by comparing different models in the same family via the Bayes factor, the ratio of the posterior probability of the model given the observations.
The authors utilize a BIC-based approximation
to the Bayes factor that is much easier to compute. 
A remarkable highlight of this framework is the simultaneity of selecting variables and the number of clusters, which overcomes the drawbacks of the sequential model selection procedure. However, the method is not applicable to the current spectral vertex clustering task in the sense that neither the distributional model nor the greedy variable selection algorithm is appropriate with respect to the graph context. This inspires the development of a reliable model for spectral embedding encompassing both signal and noise dimensions and a novel methodology for vertex clustering on graphs with heterogeneous community structure. Note that our spectral graph clustering setting allows the use of the standard BIC rather than
the Bayes factor approximation used in \cite{raftery2006variable}.

A simultaneously-developed, related approach to ours, using a Bayesian modeling perspective, is presented in \cite{heard2019}. The basic
model utilized is the same as in this paper, with the extra complexity of the 
prior distributions on the parameters and their associated hyper-parameters. These are fit using Markov Chain
Monte Carlo, rather than the frequentist perspective presented herein. One advantage of the
frequentist approach is the relative ease in handling very large graphs.
We also propose a 2-step procedure
which, while it does not produce a maximum likelihood solution, seems empirically 
to perform as well as the full maximum likelihood
procedure.
Of course Bayesian inference has its advantages (e.g., uncertainty quantification) and variational methods allow for scalable Bayesian inference.
Still, \cite{heard2019} use our basic distributional model, and we provide an extensive simulation analysis in Section 3.2 to justify our enabling conjecture.

The paper is organized as follows. In Section \ref{sec: Random graphs}, we review the existing methodology of GMM $\circ$ ASE, that is, adjacency spectral embedding followed by Gaussian mixture modeling. In Section \ref{sec: Models for Extended ASE}, we define an extension of adjacency spectral embedding and provide a specific Gaussian mixture model to characterize the potential distribution of the extended ASE based on our simulation results. In Section \ref{sec: Simultaneous Model Selection}, we propose a simultaneous model selection framework, as well as two heuristic algorithms specifically tailored for graphs under a stochastic block model. In Section \ref{sec: Experimental Results}, simulation results and an illustrative real data application are presented. We conclude the paper by remarking in Section \ref{sec: Conclusion} on some extensions of our approach.

\section{Background}
\label{sec: Random graphs}
In this section we review the ubiquitous spectral graph clustering with sequential model selection. We first introduce models for random graphs, then we summarize the existing methodology.

\subsection{Random dot product graph and stochastic block model}
One of the simplest generative models for a random graph is the \textit{inhomogeneous Erd\H os-R\' enyi} (IER) graph \citep{erdos1960evolution} on $n$ vertices, where the edges are independent Bernoulli random variables each with their own probabilities given by an $n \times n$ matrix
$P$ called the \textit{edge probability matrix}. 
To more effectively model low-rank heterogeneity, we consider the so-called \textit{random dot product graph} (RDPG) \citep{young2007random,athreya2017statistical}, one instance of the well-studied class of \textit{latent position graphs} \citep{hoff2002latent}. The definition of RDPG is as follows:

\begin{definition}[Random dot product graph (RDPG)]
    Let $\mathcal{X} \subset \mathbb{R}^d$ be a subset of $\mathbb{R}^d$ satisfying $x^T y \in [0,1]$ for all $x,y \in \mathcal{X}$. Let $X_1, \dots, X_n \in \mathcal{X}$ be $n$ latent vectors, and $X \in \mathbb{R}^{n \times d}$ be the \textit{latent position matrix} such that the $i$th row of $X$ is $X_i^T$. If the edges of an undirected graph $G$ are generated according to an edge probability matrix $P = XX^T$, then we say $G$ is a \textit{random dot product graph} (RDPG) with latent position matrix $X$, denoted by $G \sim \text{RDPG}(X)$. That is, $A_{ij}$, the entry of the adjacency matrix $A$, is independently Bernoulli distributed with parameter $P_{ij} = X_i^T X_j$, i.e.
    \begin{equation}
        \mathbb{P}[A_{ij}] = (X_i^T X_j)^{A_{ij}} (1-X_i^T X_j)^{1-A_{ij}}
    \end{equation}
    for all $(i,j) \in [n] \times [n]$.
    \label{def: RDPG}
\end{definition}
In the RDPG, the probability of the connection between vertex $i$ and $j$, namely the $i,j$-th entry of the edge probability matrix $P$, depends on the inner product of the corresponding latent positions  \citep{ hoff2002latent,hoff2005,hoff2008}. In the context of vertex clustering, vertices from the same group are supposed to share common connection attributes. Therefore, we may further assume vertices from the same group have the same latent position. This leads to the \textit{stochastic block model} (SBM) \citep{holland1983stochastic}, in which the set of vertices is partitioned into $K$ groups, called blocks. The connectivity of the graph is parameterized by the \textit{block connectivity probability matrix} $B$, which determines the edge probability within and between blocks. The formal definition of SBM is given below:
\begin{definition}[Stochastic block model (SBM)]
    Let $G$ be the graph of interest with $n$ vertices, $B \in [0,1]^{K \times K}$ be the block connectivity probability matrix, and $\bm\pi = (\pi^{(1)}, \dots, \pi^{(K)}) \in (0,1)^{K}$ be the vector of block membership probabilities such that $\sum_{k=1}^K \pi^{(k)} = 1$. $G$ is called a $K$-block \textit{stochastic block model} (SBM) graph, denoted by SBM$(n,B,\bm\pi)$, if there is a random vector $\bm\tau = (\tau_1, \dots, \tau_n)$, called the \textit{block memberships}, that assigns vertex $i$ to block $k$ with probability $\pi^{(k)}$. Mathematically, $\tau_1, \dots, \tau_n$ are i.i.d.\ categorical random variables distributed according to parameter $\bm\pi$, i.e.
    \begin{equation}
        \mathbb{P}[\tau_i = k] = \pi^{(k)}
        \label{eq: distribution of block membership}
    \end{equation}
    for all $i \in [n]$ and $k \in [K]$. Furthermore, the edges are generated according to an edge probability matrix $P$, whose $i,j$-th entry equals $B_{\tau_i, \tau_j}$. Equivalently, $A_{ij}$, the entry of the adjacency matrix $A$, is independently Bernoulli distributed with parameter $P_{ij} = B_{\tau_i, \tau_j}$, i.e.
    \begin{equation}
        \mathbb{P}[A_{ij}] = (B_{\tau_i, \tau_j})^{A_{ij}} (1-B_{\tau_i, \tau_j})^{1-A_{ij}}
    \end{equation}
    for all $(i,j) \in [n] \times [n]$.
    \label{def: SBM}
\end{definition}

In addition, it is convenient in some cases to consider that the block membership vector $\tau$ is not random but fixed. We call such a graph a \textit{stochastic block model conditioned on block memberships}, denoted by SBM$(B, \bm\tau)$. If an undirected graph $G \sim \text{SBM}(B, \bm\tau)$ and $B$ is positive semidefinite, then $G$ can be represented by an RDPG, the random dot product graph, with at most $K$ distinct latent positions. In this case, all vertices in the same block have the same latent vectors. This provides the connection between the SBM with positive semi-definite block connectivity probability matrix and the RDPG. For the relationship between an SBM with indefinite block connectivity probability matrix and a generalized RDPG, we refer the reader to \cite{rubin2017statistical}.

\subsection{Spectral graph clustering via adjacency spectral embedding}
Given an observed stochastic block model graph, our inference task is to identify the underlying memberships of the vertices corresponding to the blocks to which they belong. That is, if $G \sim \text{SBM}(B,\bm\tau)$, our goal is to infer the graph parameter $\tau$ (in the presence of nuisance parameters $K$ and $B$) from the observed adjacency matrix $A$.
Among various techniques, spectral clustering methods \citep{von2007tutorial} are effective, well-studied and computationally feasible approaches through which the vertices of a graph are mapped to points in Euclidean space. These Euclidean points are the data on which traditional clustering methods can be applied to finalize the clustering procedure. Spectral clustering performs spectral decomposition on some ``similarity'' matrix that represents the graph. There are two natural similarity matrices, namely the adjacency matrix and the Laplacian matrix of the graph. While the choice between adjacency matrix and Laplacian matrix is always debatable, it has been shown that neither of them dominates the other \citep{tang2018limit,CAPENETSCI,priebe2018two}. 
Properties of the top eigenvectors of the adjacency matrix, known as the \textit{adjacency spectral embedding} (ASE), have been analyzed in the literature
(\cite{sussman2012consistent, sussman2014consistent, athreya2016limit,lyzinski2017community};
see \cite{athreya2017statistical} for a recent survey)
where it has been proven that the rows of ASE converge to the corresponding underlying latent positions.
In this paper, we focus on the spectral method using the adjacency matrix for ease of analysis.

There are two model selection problems in spectral clustering of an SBM graph. One is to estimate the embedding dimension $d$, while the other is to estimate the number of blocks $K$. The traditional solution of these two model selection problems proceeds sequentially, namely applying variable selection or dimension reduction techniques to estimate $d$ (let the estimate be $\hat{d}$) first, then applying model selection techniques on the data with $\hat{d}$-dimensional adjacency spectral embedding to estimate $K$. Since the two model selection procedures are executed in sequence, we refer to this approach as \textit{sequential model selection}.
For example,
one can apply any so so-called \textit{scree plot method} (an effective method to locate the ``elbow'' in the scree plot is proposed in \cite{zhu2006automatic}; we refer to this method as ``ZG" in this paper) to estimate the embedding dimension, and then apply the BIC approach \citep{keribin2000consistent} on the spectral embedding to select the number of clusters for the subsequent GMM clustering. We denote this sequential model selection approach BIC $\circ$ ZG for the purpose of comparison.

\section{Models for Extended Adjacency Spectral Embedding}
\label{sec: Models for Extended ASE}

\subsection{Extended adjacency spectral embedding}
In practical settings, the rank of the edge probability matrix $P$, namely the ideal embedding dimension $d$, is unknown, because $P$ is unobserved and we observe only the adjacency matrix $A$ which is a noisy version of $P$. To address the problem of estimating $d$, we hereby define the \textit{extended adjacency spectral embedding} (extended ASE):
\begin{definition}[Extended adjacency spectral embedding (extended ASE)]
    Let $G$ be an undirected graph with $n$ vertices, and $A \in \mathbb{R}^{n \times n}$ be its symmetric adjacency matrix. Let the spectral decomposition of $A$ be
    \begin{equation}
        A = \hat{U} \hat{\Lambda} \hat{U}^T
        \label{eq: spectrual decomposition of adjacency matrix}
    \end{equation}
    Here, $\hat{\Lambda} \in \mathbb{R}^{n \times n}$ is a diagonal matrix with eigenvalues of $A$ on its diagonal in descending order. That is, $\hat{\Lambda} = \text{diag}(\hat\lambda_1, \dots, \hat\lambda_n)$ with $\hat\lambda_1 \ge \cdots \ge \hat\lambda_n$. $\hat{U}$ is an orthogonal matrix whose columns are the corresponding eigenvectors of $A$. For a given integer $D$ satisfying $1 \le D \le n$, called the \textit{embedding dimension}, the \textit{extended adjacency spectral embedding} (extended ASE) of $G$ with dimension $D$ is given by
    \begin{equation}
        \hat{Z} = \hat{U}_{[D]} \hat{\Lambda}_{[D]}^{\frac{1}{2}}
        \label{eq: extended ASE}
    \end{equation}
    where $\hat{U}_{[D]} \in \mathbb{R}^{n \times D}$ is the submatrix of $\hat{U}$ consisting of its first $D$ columns, and $\hat{\Lambda}_{[D]} \in \mathbb{R}^{D \times D}$ is the submatrix of $\hat{\Lambda}$ consisting of its first $D$ rows and columns.
    \label{def: extended ASE}
\end{definition}
In practice, $D$ can be taken as a loose upper bound for $d$. Because the methodology developed herein allows $D \gg d$ and allows subsequent post-embedding estimation of $d$, such an upper bound is usually easily obtained
either from first principle assumptions about the problem
or other external information. 
Typically one considers the scree plot---the plot of the decreasing eigenvalues against $d$---and looks for an ``elbow'' using a profile likelihood \citep{zhu2006automatic} or similar
method. Since the number of clusters $K$ is a bound on the rank of $P$, one could use the 
maximum value of $K$ as the choice for $D$. Alternatively, one can look at scatter plots of the embedding to look for dimensions
in which no clustering is apparent; this is perilous, but can provide some information to
add to that of scree plots or other information. Finally, if one has a Bayesian inclination, and
can suggest a reasonable prior on $d$, this prior can be used to determine a reasonable choice
for $D$.
The number of blocks gives an upper bounded for the rank of $P$, but since this is unavailable, we must rely on other methods. Once again, the information that would be used in a Bayesian prior on
the number of blocks can be used to determine an upper bound on $K$.
A great advantage of our approach is that we are happy with (perhaps greatly) over-estimating $d$ with $D$,
as our methodology allows this first choice to be remedied later, whereas conventional spectral clustering is constrained to proceed with the first embedding dimension.

In this paper, we assume $D$, the embedding dimension of the extended ASE, is always given without estimation. 
From the formula (\ref{eq: extended ASE}), it is immediate that the first $d$ columns of $\hat{Z}$ is the regular ASE, the adjacency spectral embedding, $\hat{X}$. The extended ASE $\hat{Z} \in \mathbb{R}^{n \times D}$ can be partitioned into two parts as $\hat{Z} = \left[\hat{X} | \hat{Y} \right]$, where $\hat{X} \in \mathbb{R}^{n \times d}$ and $\hat{Y} \in \mathbb{R}^{n \times (D-d)}$. We call the first $d$ dimensions $\hat{X}$ the \textit{informative} part, while we call the remaining dimensions $\hat{Y}$ the \textit{redundant} part. If we consider the spectral decomposition of $P$, the unperturbed version of $A$, then all of the latent position information is contained in the first $d$ dimensions, justifying our terminology. We notice that all existing methods using ASE can be applied to the extended ASE simply by truncating to an estimated embedding dimension $\hat{d}$. Moreover, as our main result in the paper, the extended ASE can be used to perform simultaneous model selection and vertex clustering without first estimating the embedding dimension $d$.

\subsection{Distributional results for extended ASE}
\label{sec: Distributional results for extended ASE}
We seek a model-based clustering approach to perform vertex clustering directly on the extended ASE. In the framework of model-based clustering, both the informative part and the redundant part need to be jointly parameterized so as to make all models comparable. For this purpose, we need to provide a model for the entire extended ASE. A remarkable distributional result for the informative part of the extended ASE is available \citep{athreya2016limit, tang2018limit}. In \cite{athreya2016limit}, a central limit theorem for the rows of ASE for the RDPG, the random dot product graph, is provided. This result justifies GMM clustering for identifying the block memberships in a stochastic block model via ASE. In \cite{tang2018limit}, the central limit theorem of ASE is restated in a stronger version, in the sense that its proof does not need an assumption that has been made in \cite{athreya2016limit}. Basically, the theorem states that any row of the ASE of an RDPG asymptotically follows a multivariate normal distribution centered at its conditional latent position (up to orthogonal nonidentifiability). Specifically, let $G \sim \text{SBM}(n,B,\bm\pi)$, i.e., $G$ is a stochastic block model graph. Considering the latent positions themselves follow an i.i.d.\ categorical distribution into $K$ distinct possible $d$-dimensional vectors according to $B$, the unconditioned version of the theorem claims that any row of the ASE of the graph $G$ converges in distribution to a mixture of $K$ multivariate normals, with mixing probabilities $\bm\pi$. The theorem gives a complete formula for the covariance matrix of each multivariate normal component, thus fully characterizing the marginal distributional behavior of the informative part of the extended ASE. 

\begin{figure}[htp]
    \centering
    \subfigure[Informative vectors for a 2-block graph.]{
    \includegraphics[width=0.45\columnwidth,clip=true]{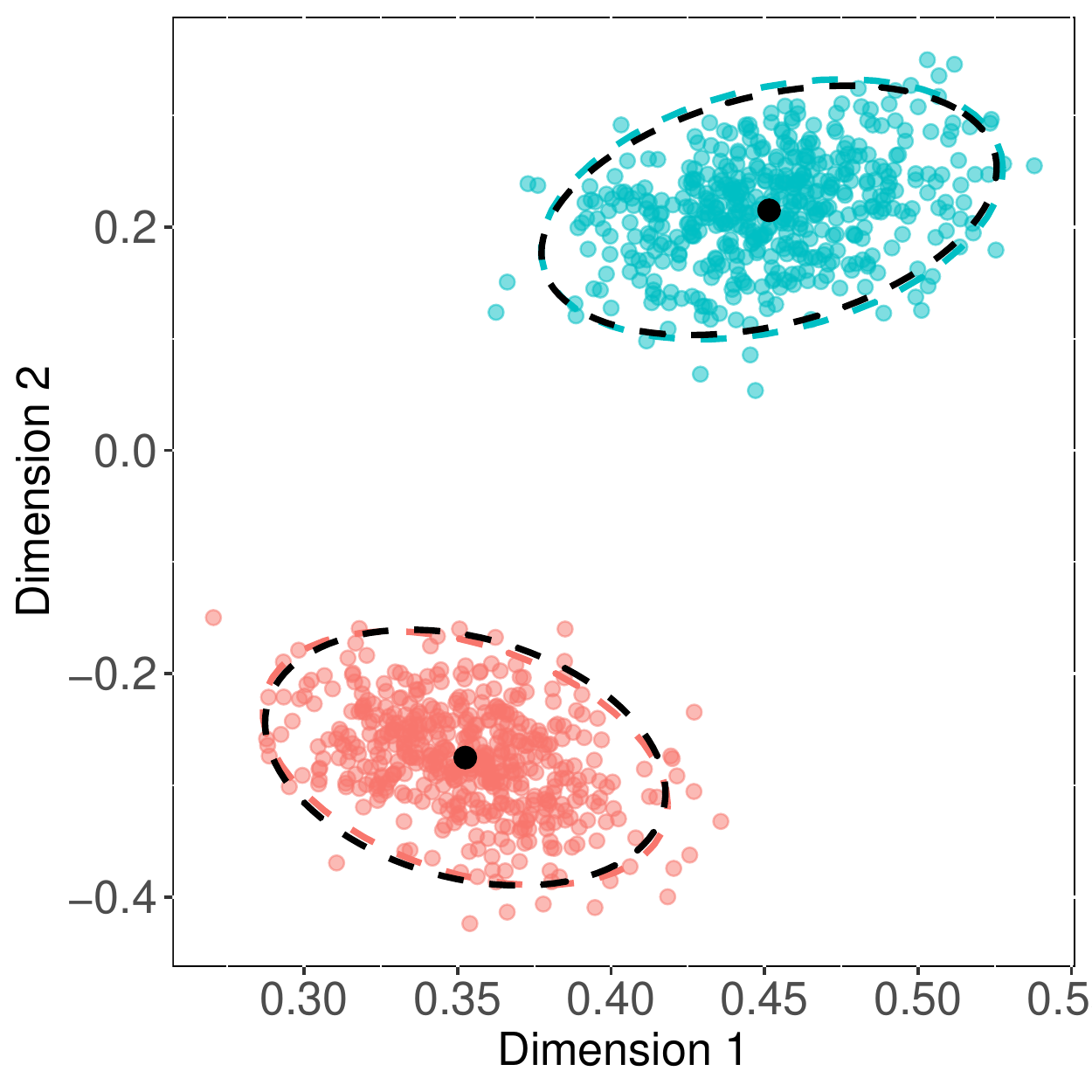}}
    \subfigure[Informative vectors for a 3-block graph.]{
    \includegraphics[width=0.45\columnwidth,clip=true]{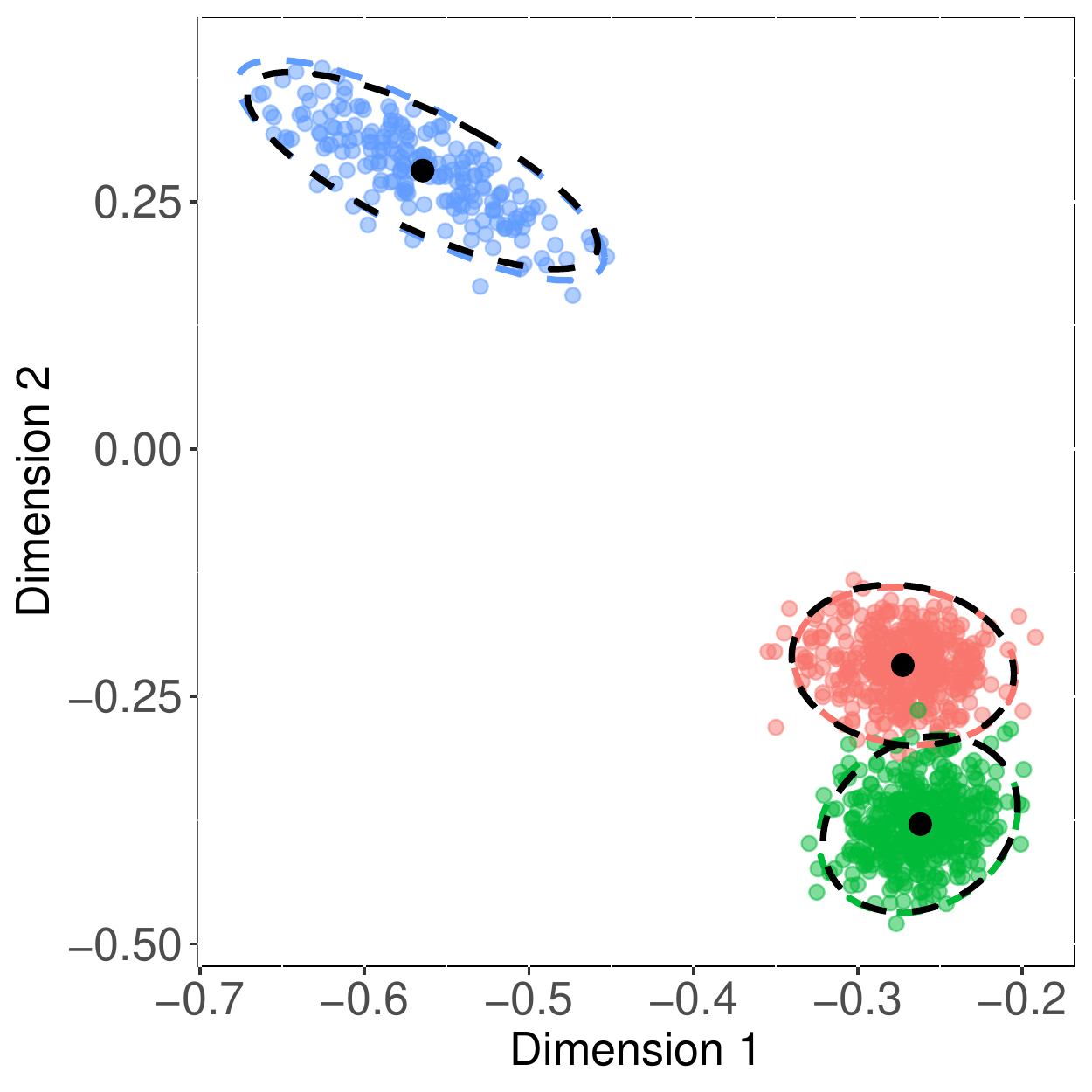}}
    \caption[The informative part $\hat{X}$.]{(a) 
    The informative part $\hat{X}$ for a graph
    drawn from the $2$-block SBM. 
    (b) The first two dimensions of the
    informative part $\hat{X}$ for a graph drawn from the 3-block SBM.
    In both cases,
    the number of vertices is $n=1000$, color denotes block
    membership, the color ellipses are the 95$\%$ level curves for the empirical distributions, the black dots represent true $X$'s, and the black ellipses are the 95$\%$ level curves for the theoretical distributions.
(The empirical and theoretical ellipses are nearly indistinguishable.)
     The SBM parameters $B$ and $\bm\pi$ for the two cases are given in the text.}
    \label{fig: ch3_inform}
\end{figure}

To obtain empirical understanding of the distributional behavior of the redundant part of extended ASE, we consider a suite of simulations. We generate random graphs according to the stochastic block model SBM$(n ,B, \bm\pi)$, where $$B =
\begin{bmatrix}
  0.2 &0.1 \\
  0.1 &0.25
\end{bmatrix}
$$ and $\bm\pi = (0.5, 0.5)$ for the $K$=$2$-block case, and $$B =
\begin{bmatrix}
  0.2 &0.1 &0.08 \\
  0.1 &0.25 &0.05 \\
  0.08 &0.05 &0.4
\end{bmatrix}
$$ and $\bm\pi = (0.4, 0.4, 0.2)$ for the $K$=$3$-block case. The number of vertices considered, $n$, is varied during the simulation study. Notice that the true embedding dimension $d=rank(B)$ is $d = 2$ for the $2$-block case and $d=3$ for the $3$-block case. We apply the extended ASE to the adjacency matrix $A$ according to Definition \ref{def: extended ASE}. As defined, the extended ASE $\hat{Z} \in \mathbb{R}^{n \times D}$ is partitioned into informative part $\hat{X} \in \mathbb{R}^{n \times d}$ and redundant part $\hat{Y} \in \mathbb{R}^{n \times (D-d)}$ by $\hat{Z} = \left[\hat{X} | \hat{Y} \right]$. $\hat{Y}_i \in \mathbb{R}^{D-d}$ denotes the $i$-th row of $\hat{Y}$, which corresponds to the $i$-th vertex with block membership $\tau_i$. 

Figure \ref{fig: ch3_inform} depicts the informative part $\hat{X}$ for the two different
block models described above, entirely in agreement with theory. Colors indicate block membership. Each
graph was drawn with $n=1000$ vertices.

The observations of the distributional behavior of the redundant part $\hat{Y}$ are as follows.

\begin{figure}[ht]
    \centering
    \subfigure[Sample mean values for $n=200$ and $n=2000$.]{
    \includegraphics[width=0.48\columnwidth,clip=true]{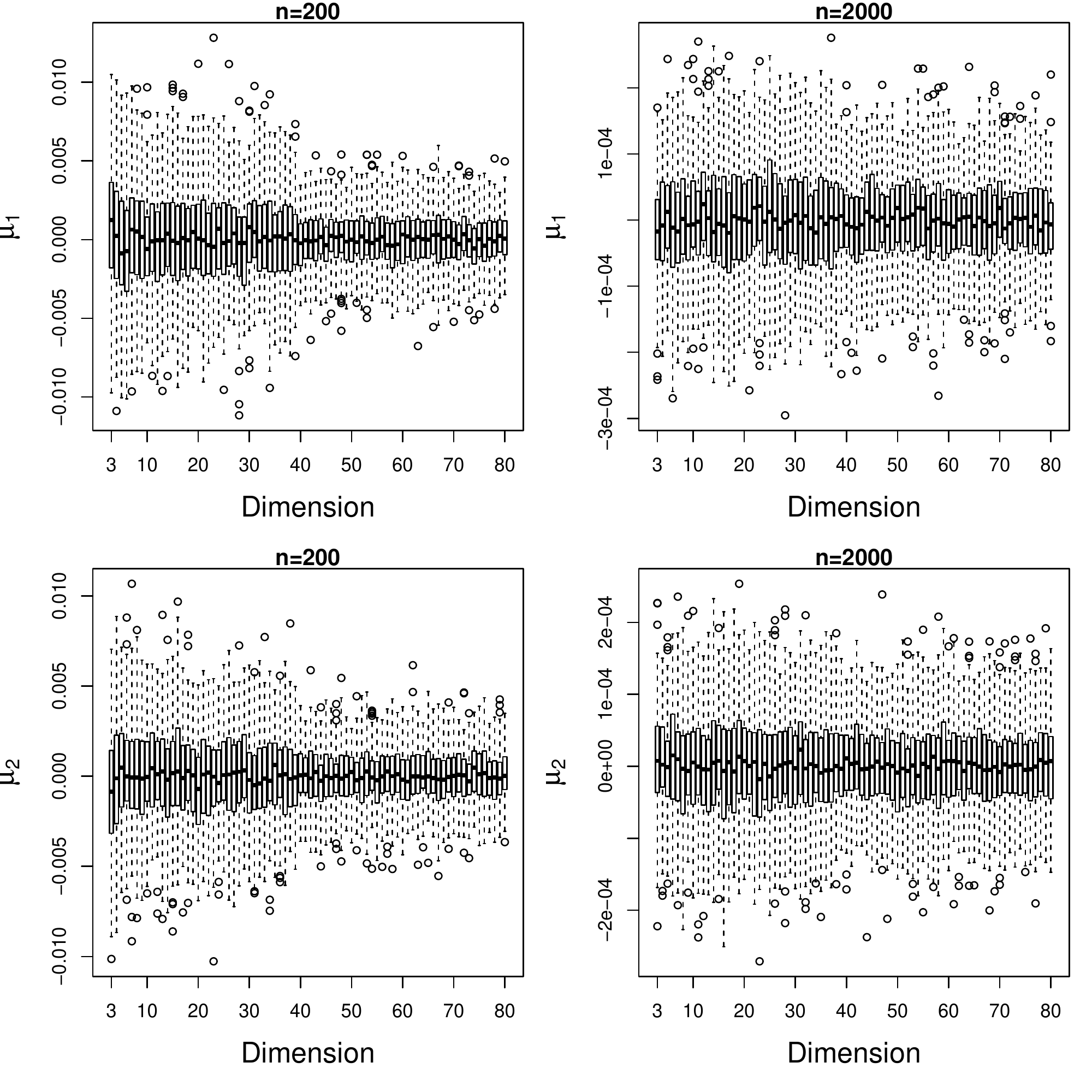}} 
    \subfigure[Sample mean values for $n=200$ to $16,000$.]{
    \includegraphics[width=0.48\columnwidth,clip=true]{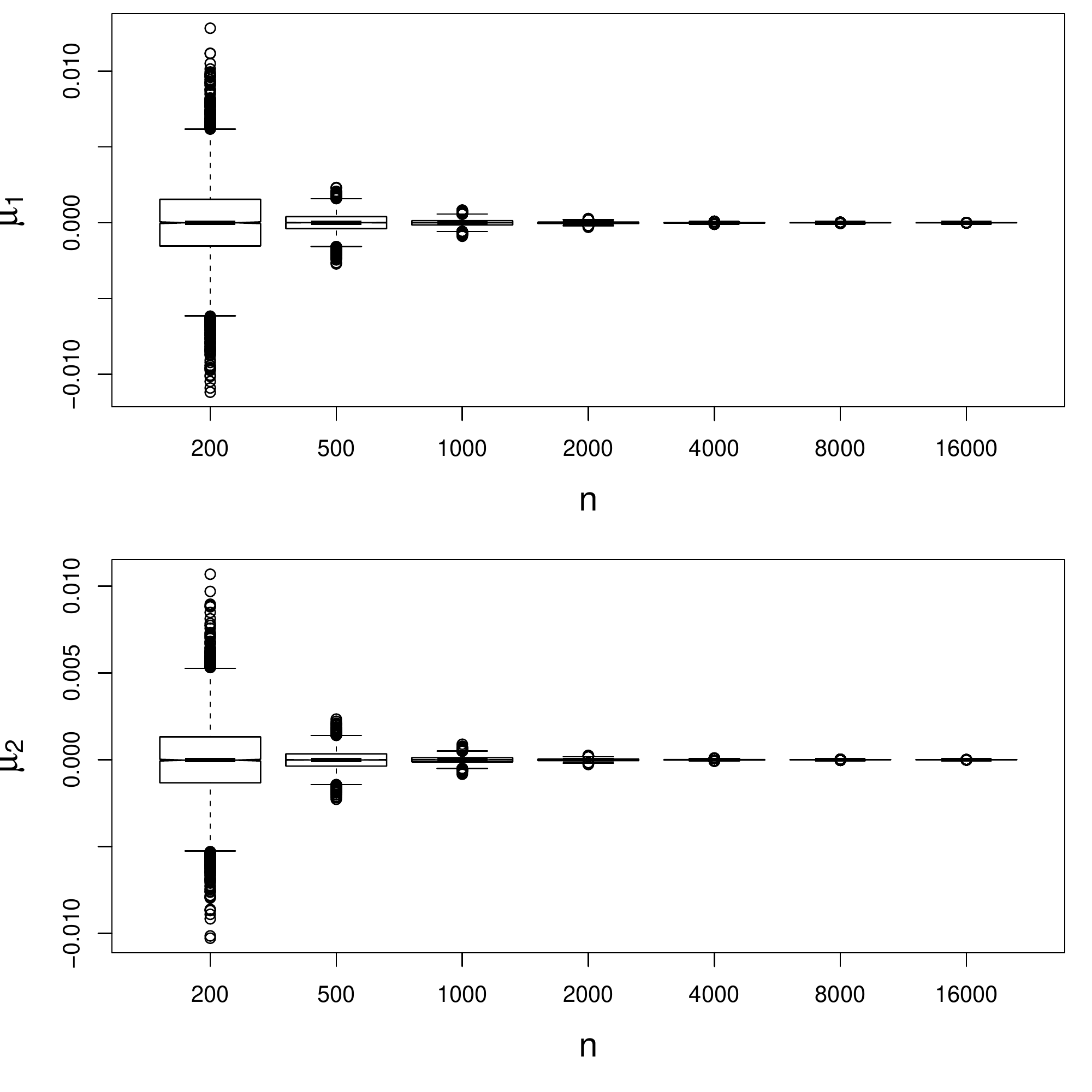}}
    \caption[The sample mean of the redundant part $\hat{Y}_i$ for all $i$ in block 1.]{(a) 
    Boxplots of $100$ Monte Carlo replicates of the sample mean of the redundant part $\hat{Y}_i$ 
    for all $i$ in block 1 (top) and block 2 (bottom). 
    The graphs are drawn from the $2$-block SBM($n$, $B$, $\bm\pi$) with $B$ and $\bm\pi$ given in the text. 
    The number of vertices of the graphs is $n=200$ (left) and $n=2000$ (right). 
    The extended ASE is applied with dimension $D = 80$. 
    (b) The sample mean of the redundant part of 
    $\hat\mu^{(k)}_1$ for $n=200$ to $n=16,000$. Again, the
    top plot corresponds to block 1, the bottom to block 2. In all the
    simulations, $100$ graphs are generated, so each box corresponds to $100$ observations.}
    \label{fig: ch3_asemean}
\end{figure}

\noindent \textbf{Observation 1: The within-block sample mean of rows of $\hat{Y}$ tends to zero as $n$ increases.} Figure \ref{fig: ch3_asemean} shows the results for the sample mean for the 2-block model. Denoted by $\hat\mu^{(k)} \in \mathbb{R}^{D-d}$, the sample mean of the rows of redundant part $\hat{Y}$, for all the rows belonging to block $k$, is calculated by $(k = 1, 2)$
\begin{equation}
    \hat\mu^{(k)} = \frac{1}{n_k} \sum_{i: \tau_i = k} \hat{Y}_i
\end{equation}
where $n_k$ is the number of vertices assigned to block $k$. We plot the sample mean values $\hat\mu^{(k)}$ ($k = 1, 2$) for each dimension from $100$ Monte Carlo replicates in Figure \ref{fig: ch3_asemean}a. For larger $n$, the points are closer to zero in general. 
The upper tick-mark for the $n=200$ plots is $0.01$, while for $n=2000$ it is
$0.0002$.
In Figure \ref{fig: ch3_asemean}b we plot the means for $100$ Monte Carlo replicates
for various values of $n$ for the first redundant vector, 
$\hat\mu^{(k)}_1$ (where the $s$-th entry of $\hat\mu^{(k)}$ is denoted by $\hat\mu^{(k)}_s$). While this proves nothing, even for this one simulation setting, the boxplots in Figure \ref{fig: ch3_asemean}b certainly support our Observation 1.

\begin{figure}[h!]
    \centering
    \subfigure[Sample variance values.]{
    \includegraphics[width=0.48\columnwidth,clip=true]{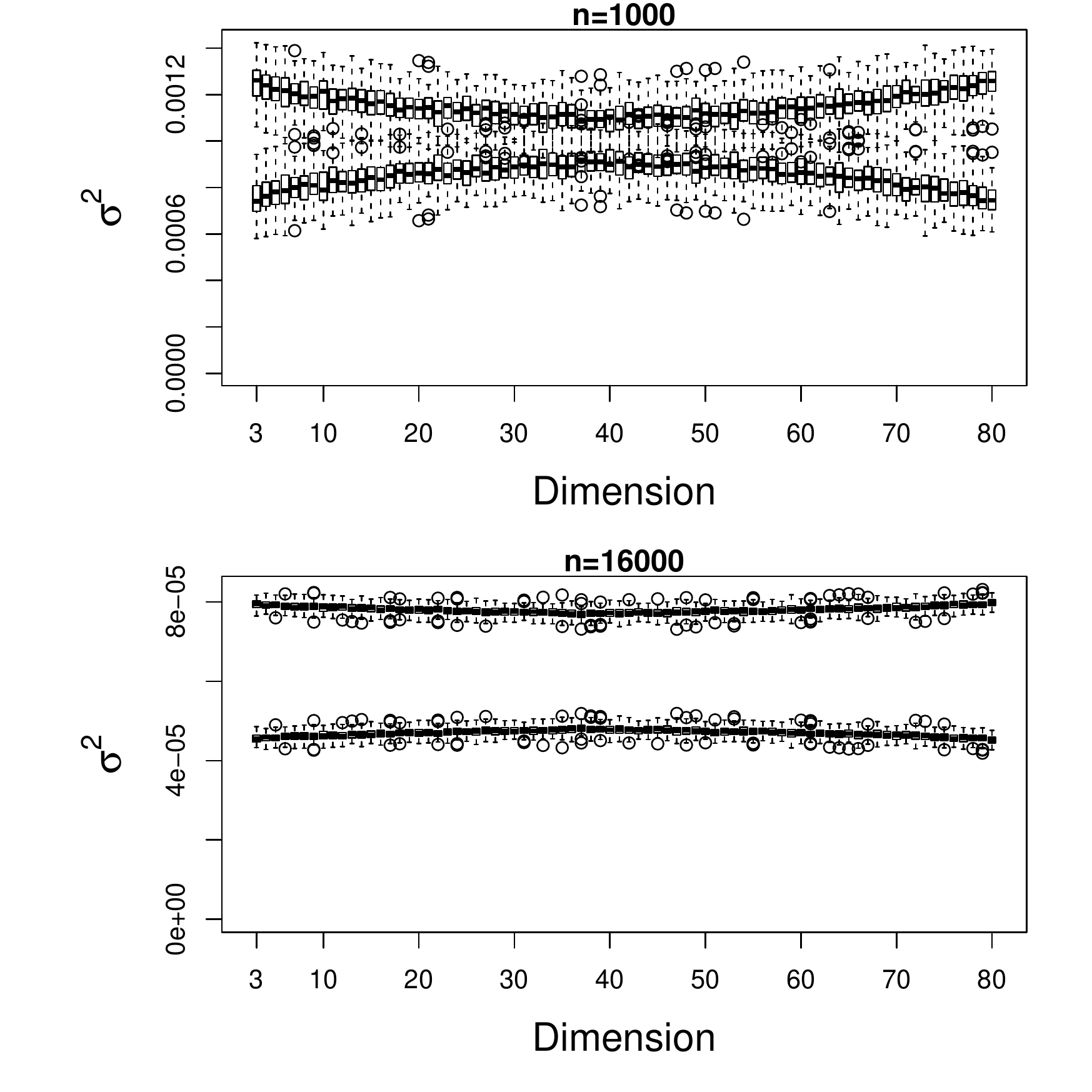}}
    \subfigure[Median curves.]{
    \includegraphics[width=0.48\columnwidth,clip=true]{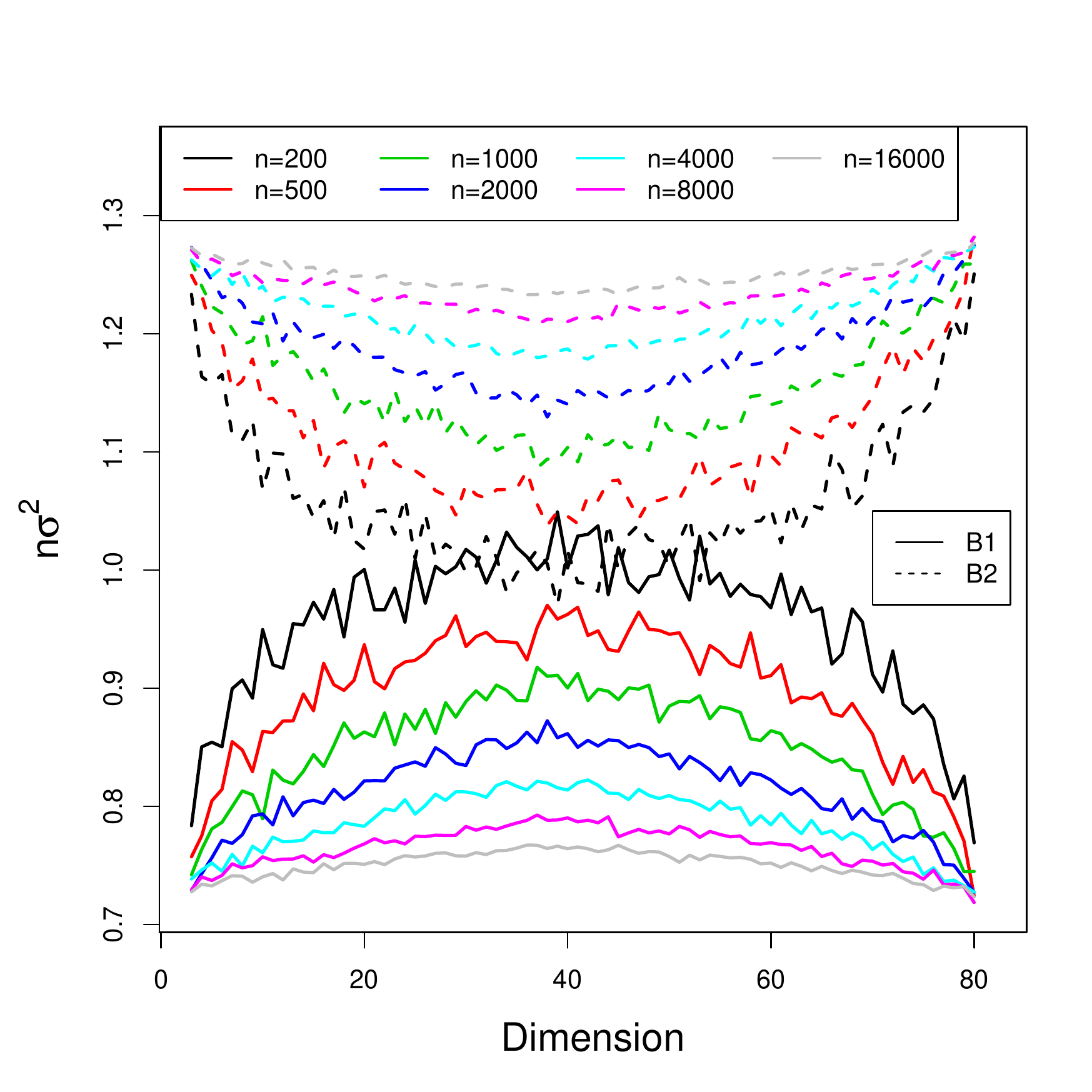}}
    \caption[The sample variance of each dimension of $\hat{Y}_i$ for all $i$.]{
    The within-block sample variance of rows of $\hat{Y}$ for each redundant dimension.
    The graphs are drawn from the $2$-block SBM($n$, $B$, $\bm\pi$) with given $B$ and $\bm\pi$. 
    The number of vertices is varied from $200$ to $16,000$, denoted by colors. 
    The extended ASE is applied with dimension $D = 80$. 
    The sample variance is calculated from dimension $3$ to dimension $80$. 
    (a) Sample variance values from $100$ Monte Carlo replicates for $n=1000$ (top) and $n=16,000$ (bottom). 
    Boxplots are used to indicate the variation of the values. In each of the two panels, the boxes with
    larger values correspond to Block 1, and the ones with smaller values correspond to Block 2.
    (b) Curves of the median of $100$ Monte Carlo replicates of the sample variance values against 
    the dimensions for the values of $n$ indicated in the legend.
    To avoid over plotting and for better comparison across the values of $n$,
    we have multiplied the value of $\hat\sigma^2$ by $n$ in this plot.
    The solid curves correspond to the first block, and the dashed curves to
    the second block.
    }
    \label{fig: ch3_asevar}
\end{figure}

\noindent \textbf{Observation 2: The within-block sample variances of rows of $\hat{Y}$ (for each dimension) tend to constants as $n$ increases, but the values of these constants are distinct across different blocks.} Figure \ref{fig: ch3_asevar} shows the results for the sample variances. For each dimension $s$, the sample variance of the rows of $\hat{Y}$, for all the rows belonging to block-$k$, is calculated by ($s = 1, \dots, D-d; k = 1, 2$)
\begin{equation}
    \hat\sigma_s^{2^{(k)}} = \frac{1}{n_k - 1} \sum_{i: \tau_i = k} \left(\hat{Y}_{i,s} - \hat\mu^{(k)}_s\right)^2
\end{equation}
where $n_k$ is the number of vertices assigned to block $k$ and $\hat\mu^{(k)}$ is corresponding sample mean in block-$k$. We plot the sample variance values $\hat\sigma^{2^{(k)}}$ for each dimension from $100$ Monte Carlo replicates in Figure \ref{fig: ch3_asevar}a.
In each panel the 
boxplots indicate the variances for the two blocks, the lower set in each panel corresponding to
the first block, the upper to the second. For relatively small graphs ($n=1000$, top panel),
the variances are clearly different across the dimensions, indicating that for
these graphs a modified model may be appropriate -- although the extra measurement
error inherent in these more complex models may argue for using the simpler 
model of a constant variance, particularly if it is assumed that there are
a small number of redundant variables. For larger graphs ($n=16,000$, bottom)
we see that the variances appear constant and distinct for the different blocks.
To investigate the structure of the variance for a range of graph orders, 
we show in Figure \ref{fig: ch3_asevar}b the medians of the variances for $100$
Monte Carlo replicates for various values of $n$.  Again we see that for small $n$
there is clearly a difference in the variances for the different dimensions, but 
this difference becomes less pronounced for larger $n$.
Again, these empirical results, narrow though they may be, certainly support our Observation 2.

\begin{figure}[ht]
    \centering
    \subfigure[n=200]{
    \includegraphics[width=0.45\columnwidth,clip=true]{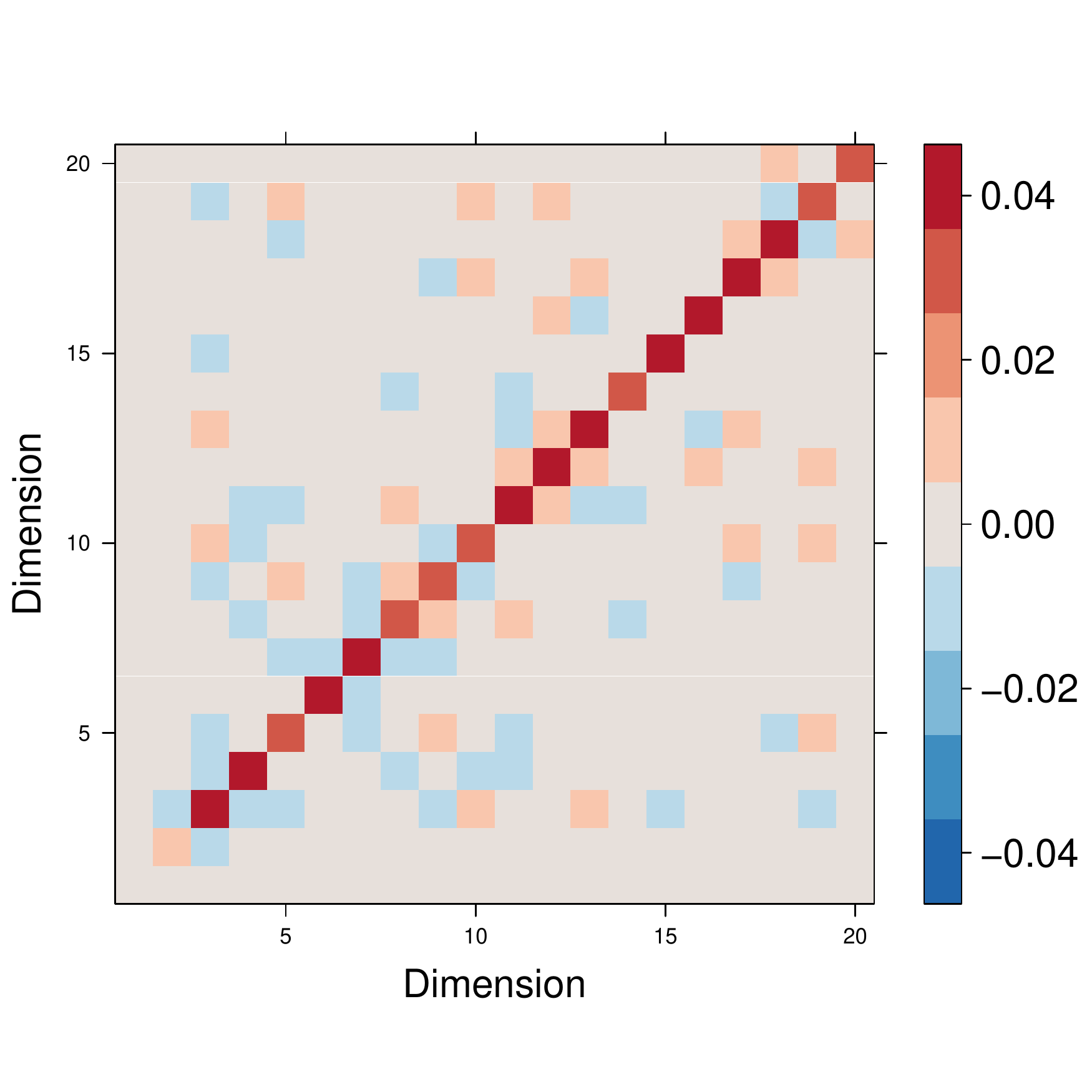}}
    \subfigure[n=2000]{
    \includegraphics[width=0.45\columnwidth,clip=true]{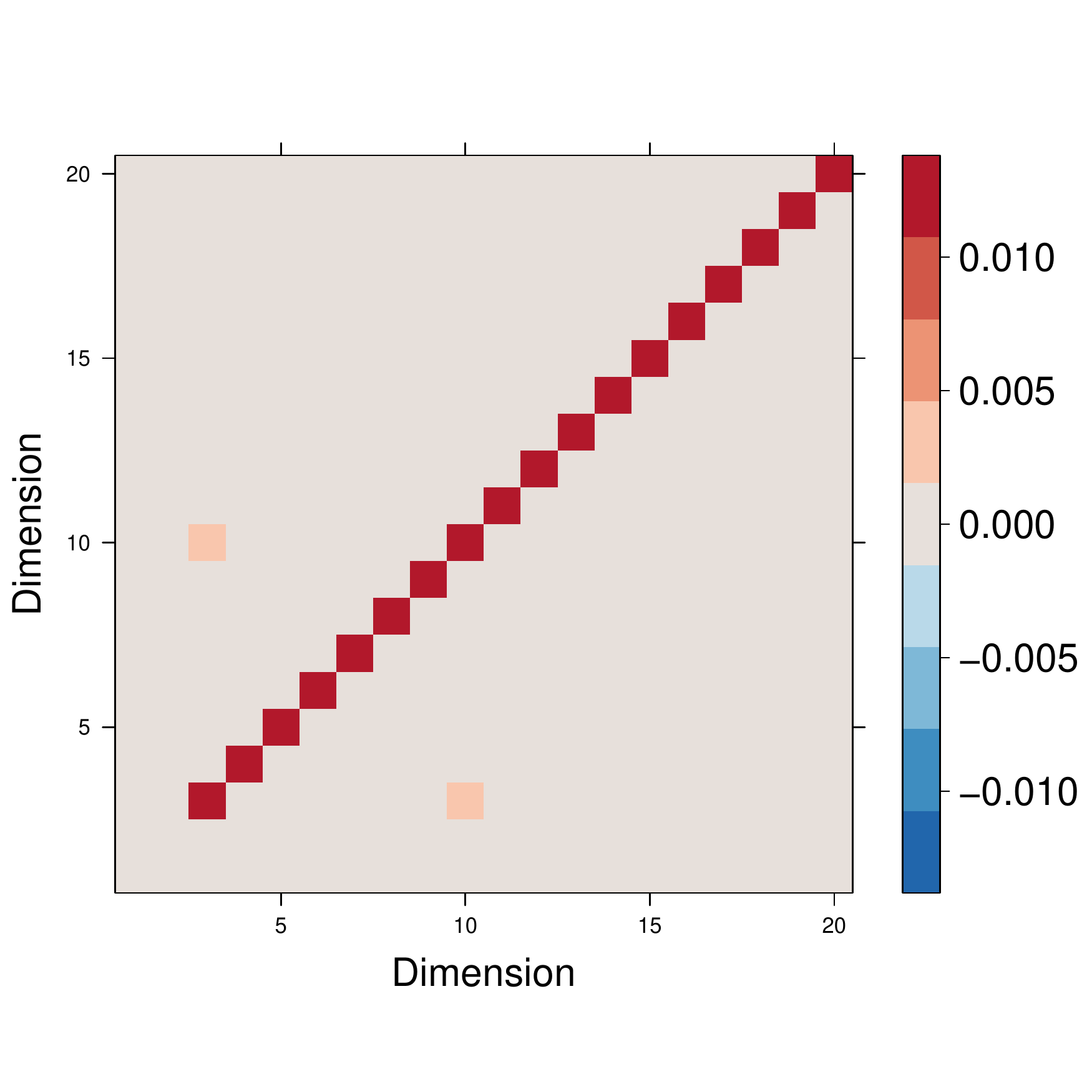}}
    \caption[The sample covariance matrix of $\hat{Z}_i$ for all vertices $i$ in block 1.]{The sample covariance matrix of $\hat{Z}_i$ for all vertices $i$ in block 1. The graphs are drawn from the $2$-block SBM($n$, $B$, $\pi$) with given $B$ and $\pi$. The extended ASE is applied with dimension $D = 20$. The x-axis and y-axis indicate the indices of the dimensions in extended ASE, respectively. Values are shown via colors. (a) $n=200$; (b) $n=2000$.}
    \label{fig: ch3_asecov}
\end{figure}

\begin{figure}[ht]
    \centering
    \subfigure[]{
    \includegraphics[width=0.45\columnwidth,clip=true]{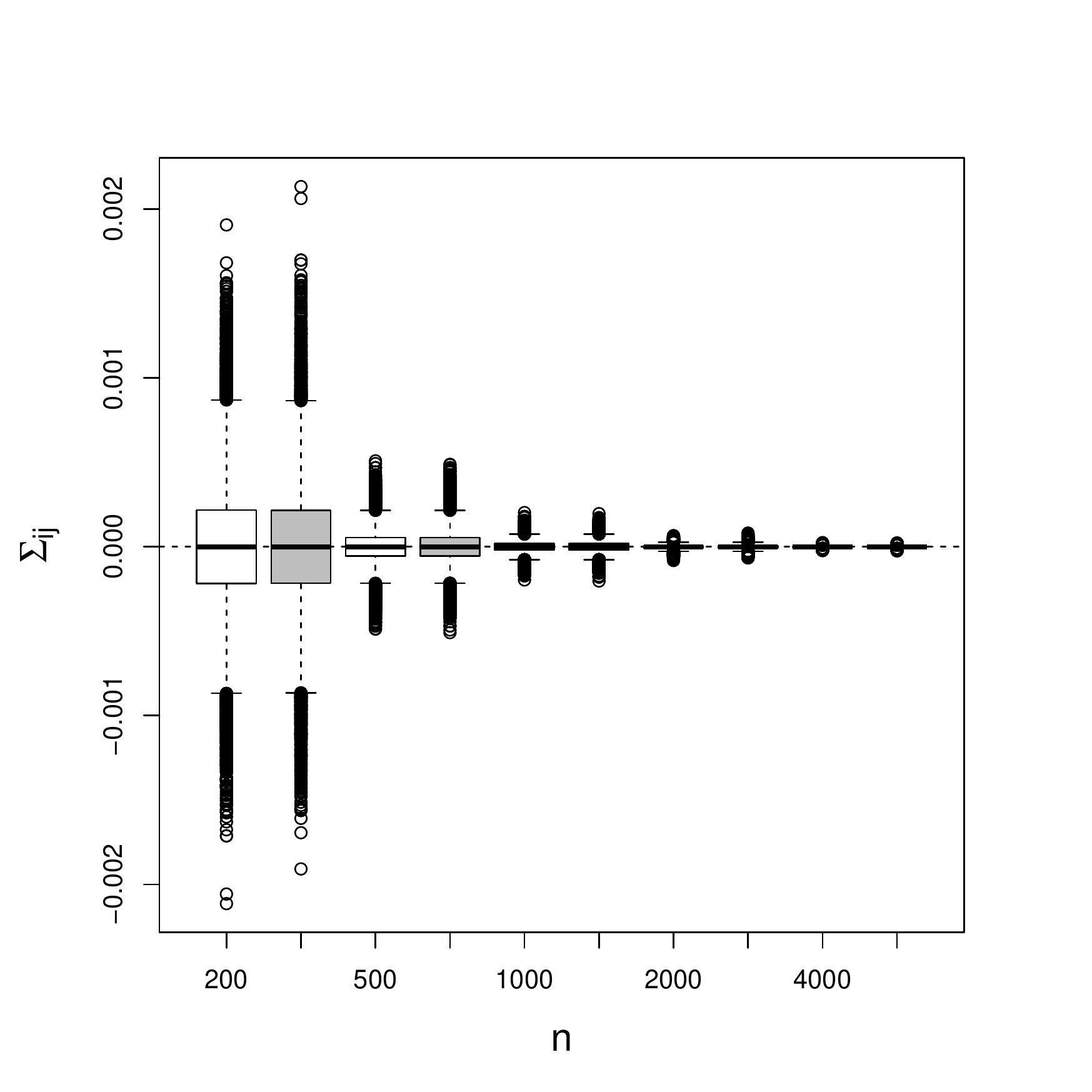}}
    \subfigure[]{
    \includegraphics[width=0.45\columnwidth,clip=true]{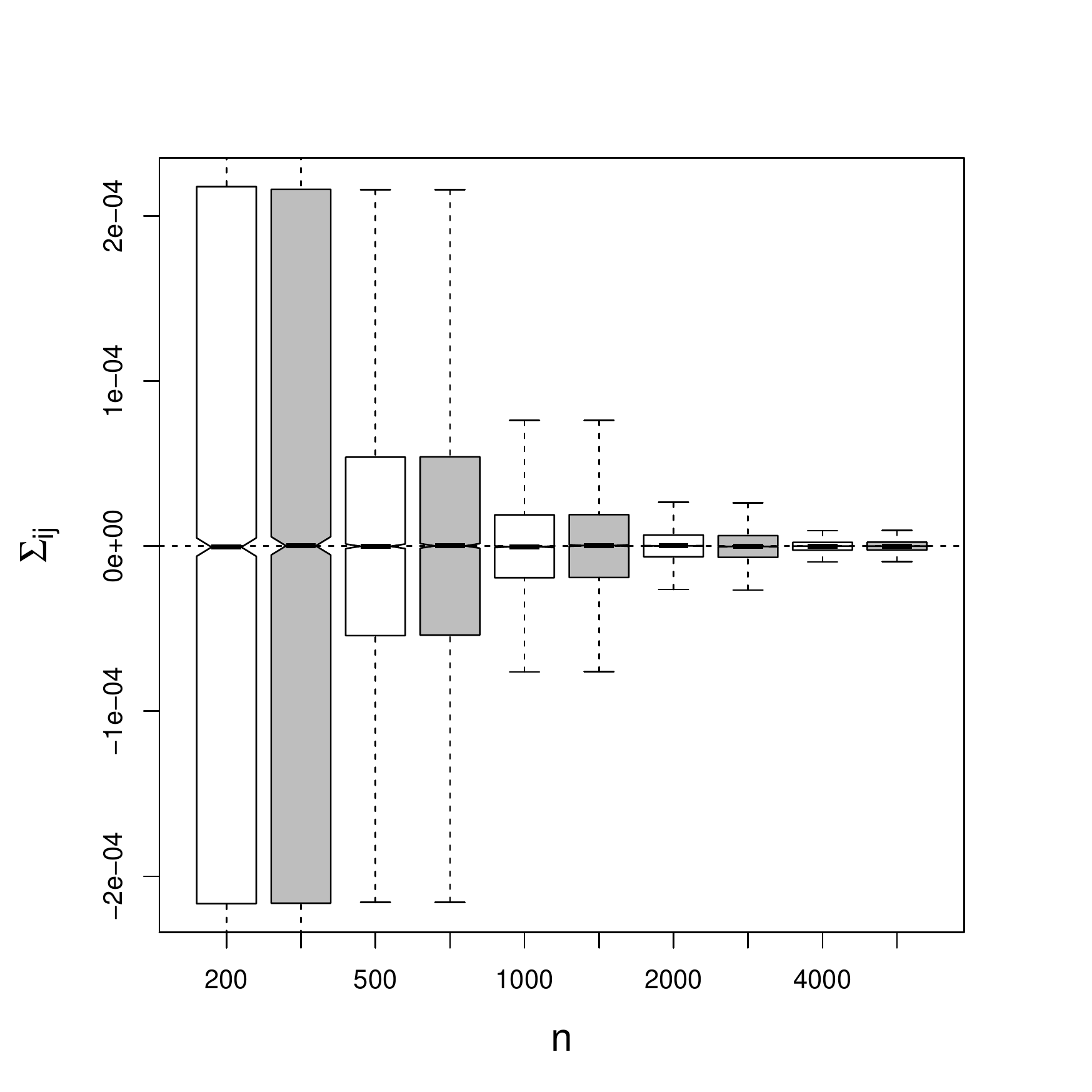}}
    \caption[The off-diagonal covariances matrix of $\hat{Y}_i$.]{
    For each value of $n$, $100$ Monte Carlo simulations are run, in which
    the off-diagonal covariances for each block is computed. Panel (a) shows the 
    boxplots of all off-diagonal terms in all the matrices; panel (b) enlarges this figure. 
    The white boxes are for block 1, the gray boxes correspond to block 2.
    }
    \label{fig: ch3_asecov2}
\end{figure}

\noindent \textbf{Observation 3: The within-block sample covariance matrix of rows of $\hat{Y}$ tends to diagonal, and the covariance between informative and redundant dimensions tends to zero, for large $n$.} Figure \ref{fig: ch3_asecov} shows the results for the sample covariance matrix. The sample covariance matrix of the rows of $\hat{Z}$, for all the rows belonging to block-$k$, is calculated by
\begin{equation}
    \hat\Sigma^{(k)} = \frac{1}{n_k - 1} \sum_{i: \tau_i = k} \left(\hat{Z}_i - \overline{\hat{Z}}^{(k)}\right) \left(\hat{Z}_i - \overline{\hat{Z}}^{(k)}\right)^T
\end{equation}
where $n_k$ is the number of vertices assigned to block $k$, $\hat{Z}_i \in \mathbb{R}^{D \times 1}$ is the $i$-th row of extended ASE (but regarded as a column vector), and $\overline{\hat{Z}}^{(k)} \in \mathbb{R}^{D \times 1}$ is the corresponding sample mean for block $k$. We plot the sample covariance matrix $\hat\Sigma^{(k)}$ for $n = 200$ in Figure \ref{fig: ch3_asecov}a and $n = 2000$ in Figure \ref{fig: ch3_asecov}b. The matrix contains both informative dimensions and redundant dimensions. We observe that the diagonal values in the matrix of redundant dimensions concentrates on a constant for $n=2000$, which is consistent with the result shown in Observation 2 and Figure \ref{fig: ch3_asevar}. The off-diagonal values in the matrix of redundant dimensions tend to zero as $n$ increases. For $n=2000$, the covariance matrix presents a block diagonal structure, partitioned by the true embedding dimension $d$. These provide evidence that the within-block sample covariance matrix of rows of $\hat{Y}$ tends to be diagonal, and the covariance between informative and redundant dimensions tends to be zero, for large $n$.
Further evidence that the off-diagonal terms are zero is given in Figure
\ref{fig: ch3_asecov2}, depicting box plots of all the off-diagonal terms 
in the redundant part for
$100$ Monte Carlo replicates for the 2-block model, for various values of $n$
between $200$ and $4000$.


The above results provide the impetus for several observations about the distribution of the extended ASE of a stochastic block model. Although we do not present results from more extensive simulations, the plot in Figure \ref{fig: ch3_inform} suggests 
(and the theoretical results mentioned support)
a GMM model with general mean vectors and covariance matrices is appropriate for the informative part of the embedding. 
For the redundant part, asymptotically, it appears that
the block-conditional means are zero, and the block-conditional covariances are diagonal.
There is some evidence that (again asymptotically) the block-conditional variances for
redundant are the same, although for small $n$, and in particularly for large $d$, these
variances may well be different, as suggested in Figure \ref{fig: ch3_asevar}b. 
For our simulations, ``small $n$'' seems to be in the thousands, but of course this would depend on the specific structure of the
stochastic block model. 
Since the extra block-conditional information in these variance
differences is likely to be small, we   will assume the simpler model where all the redundant
variances are equal within blocks, but this is an area for future investigation.

\begin{figure}[htbp]
    \centering
    \includegraphics[width=0.85\columnwidth,clip=true]{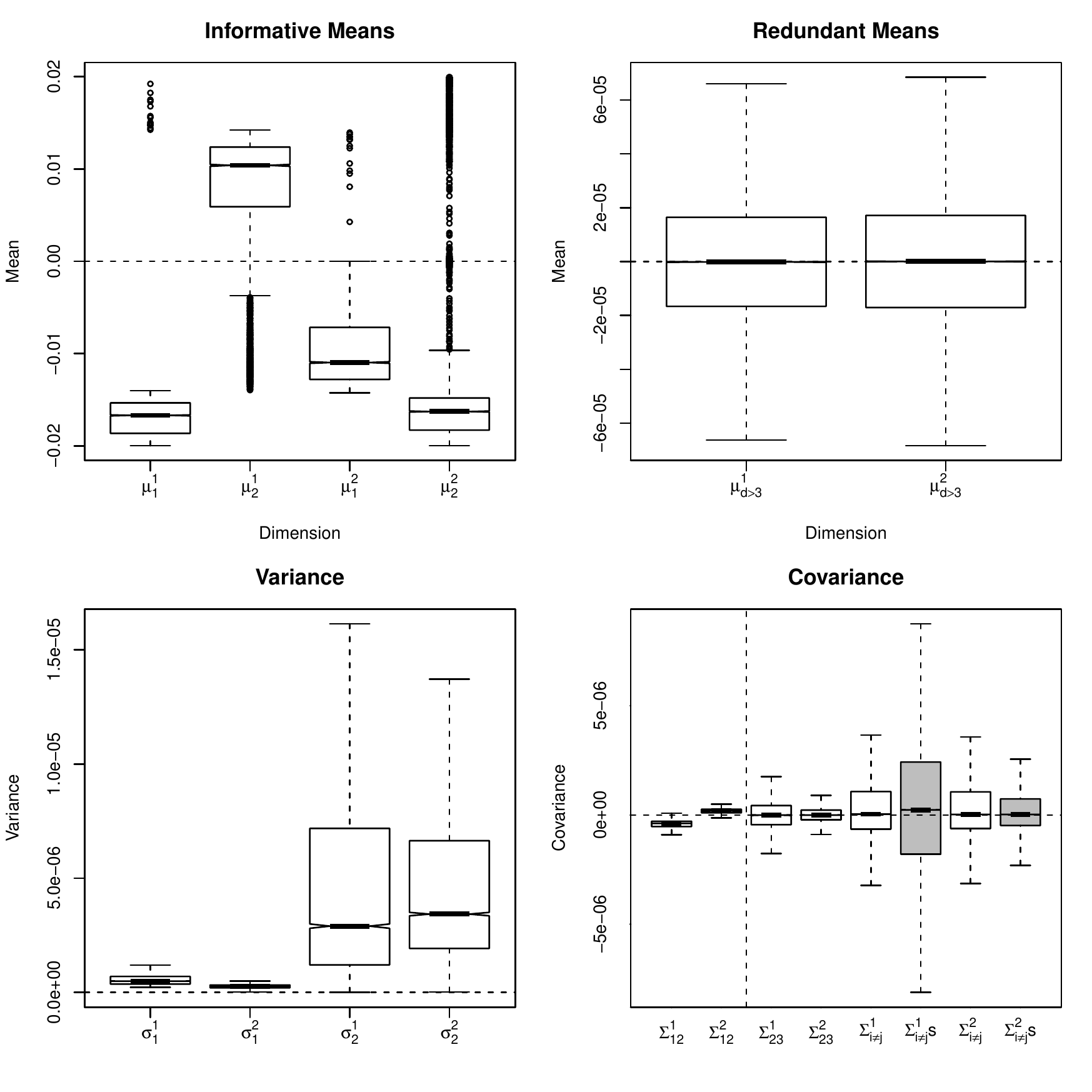}
    \caption[Large scale simulations of $\hat{Y}$.]{
    For $n=5000$, $10,000$ random symmetric $2\times 2$ probability matrices are computed
    (see the text) and
    a stochastic block model with $n_1=2500$ and $n_2=2500$
    nodes is drawn,
    and the embedding to $D=20$ is computed. 
    In all plots, the superscript corresponds to block (community) and subscripts denote
    dimension (variate).
    The top left plot shows the means for the two blocks for dimensions
    one and two. The top right plot shows the means for the two blocks for dimensions
    $s>2$. The variances for $s=1,2$ are shown in the bottom left, first two boxes,
    and for $s>2$ in the other two blocks.
    The off-diagonal covariances are shown in the bottom right plot. The two to the left
    of the dotted vertical line are for the informative dimensions, $s=1,2$, the others
    are for $s=2,3$ and $s>2$. The gray boxes show the covariance for Gaussian variates
    drawn from the corresponding model. See the text.
    }
    \label{fig: ch3_bigsims}
\end{figure}

The above simulations illustrate well the reasons for our assumptions about the model.
In order to see that these assumptions are not unique to the specific cases we consider,
we ran a much more extensive set of simulations. Just one representative example is depicted in
Figure \ref{fig: ch3_bigsims}. In this model we draw a random probability matrix 
$$
P=\left(\begin{array}{cc}
  p_1  &  q\\
  q  & p_2
\end{array}\right)
$$
with $q\sim U[0,.1]$ and $p_i\sim U[q,.2]$, ordered so that $p_1\geq p_2$.

If our observations are correct, we should expect that the
first two means are arbitrary but different, and that the $\Sigma_{s,s}$ are different for $s=1,2$.
We also expect to see $\mu_s=0$ for $s>2$, and off-diagonal covariances are $0$ for $s_1,s_2>2$.
For comparison, the gray boxes in the bottom right are a simulation in which for each graph and each block, $5000$ independent normal variates are drawn from an $18$-dimensional Gaussian with mean $0$ 
and diagonal covariance equal to the diagonal of the covariance for that block. All off-diagonal terms in the covariance fitted to the data are then collected into the gray box. This shows that the range of variation that we observe in our model is well within that which we would expect for data drawn from the model.

\subsection{Probability models for extended ASE}
For the extended ASE of an SBM graph, the theoretical result for its informative part and the conjecture for its redundant part are discussed above. There are few theoretical results about the redundant part in the literature. However, a distributional model is important and necessary for model-based clustering. So we provide a finite mixture model for the extended ASE. Although it has not been proven analytically at this point, we believe the model is asymptotically close to the truth---wrong but potentially useful, in George Box's aphorism---based on both our observations of the large sample behavior of the redundant part and the performance based on this model on the subsequent inference task.

We first state mathematically our conjectures regarding the distribution of the redundant part $\hat{Y}$ of the extended ASE. 
We consider a $K$-block SBM$(n,B,\bm\pi)$. Any row of $\hat{Y}$ is asymptotically multivariate Gaussian distributed conditioned on its block membership. That is, for any $i \in [n]$,
\begin{equation}
    \hat{Y}_i | \tau_i = k \approx N(\mu^{(k)}, \Sigma^{(k)})
\end{equation}
 if $n$ is sufficiently large. If we consider the sample statistics from the simulations to be a good estimation of the Gaussian parameters, we can further specify the model. By Observation 1, we may assume $\mu^{(k)} = 0$ for all $k \in [K]$; by Observation 2 and Observation 3, we assume $\Sigma^{(k)} = \hat\sigma^{2^{(k)}} I$, where $I$ is the identity matrix. So now our conjecture becomes
\begin{equation}
    \hat{Y}_i | \tau_i = k \approx N(0, \hat\sigma^{2^{(k)}} I).
    \label{eq: conjecture}
\end{equation}

By combining this conjecture with the theoretical results for the informative dimensions, we propose a \textit{Gaussian mixture model} (GMM) as follows. 

\begin{model}[GMM for extended ASE of undirected graphs]
    \label{mod: GMM for undirected graphs}
    Let
    \begin{equation}
        \label{eq: our model}
        f(\cdot ; \theta(d, K)) = \sum_{k=1}^{K} \pi^{(k)} \varphi(\cdot; \mu^{(k)}, \Sigma^{(k)})
    \end{equation}
    be a family of density functions for a $D$ dimensional GMM random vector, where $\{\pi^{(k)}\}_{k=1}^K$ are the mixing probabilities, $\{\mu^{(k)}\}_{k=1}^K$ are the mean vectors, and $\{\Sigma^{(k)}\}_{k=1}^K$ are the covariance matrices. Furthermore, these parameters satisfy
    \begin{equation}
        \label{eq: pi in our model}
        \sum_{k=1}^K \pi^{(k)} = 1
    \end{equation}
    \begin{equation}
        \label{eq: mu in our model}
        \mathbf{\mu}^{(k)} = [\mu_1^{(k)}, \dots , \mu_{d}^{(k)}, 0, \dots , 0]^{T}
    \end{equation}
    and
    \begin{equation}
        \label{eq: sigma in our model}
        \Sigma^{(k)} =
        \begin{bmatrix}
            \tilde{\Sigma}^{(k)} &0 \\
            0                    &\sigma^{2^{(k)}}I
        \end{bmatrix}
    \end{equation}
    where $\tilde{\Sigma}^{(k)}$ is a $d \times d$ positive semidefinite matrix, and $I$ is a $(D - d) \times (D - d)$ identity matrix. In this notation, $\theta(d, K)$ denotes the parameters $\{\pi^{(k)}, \mathbf{\mu}^{(k)}, \Sigma^{(k)}\}_{k = 1}^{K}$, specifically $\theta(d, K) = \left\{\pi^{(k)}, \left[\mu_1^{(k)}, \dots , \mu_{d}^{(k)}\right], \tilde{\Sigma}^{(k)}, \sigma^{2^{(k)}} \right\}_{k = 1}^{K}$, which belongs to the parameter space $\Theta(d, K)$
    for GMM model ${\cal F}_{d,K}$.
\end{model}

Our GMMs ${\cal F}_{d,K}$ are models for data of a given dimensionality $D$
and posit a mixture of $K$ $D$-variate normals for that data.
Because we are claiming consistency only for
the model complexity parameter estimates $d$ and $K$ --
that is,
our simultaneous model selection yields
  $(\hat{d},\hat{K}) \longrightarrow (d_0,K_0)$ as per Theorem 1 below --
and are making no claims about the parameters of the GMM itself --
the parameters of the individual multivariate normals themselves, and the mixing coefficients --
nonidentifiability within a model ${\cal F}_{d,K}$ does not concern us;
we just need identifiability in terms of the model complexity parameters $d$ and $K$.

\noindent
{\bf Identifiability Assumption.}
{\it For any $(d,K) \neq (d',K')$
there does not exist
$f \in {\cal F}_{d,K}$
and
$f' \in {\cal F}_{d',K'}$
such that $f = f'$.
This is ensured by noting that 
  none of the informative dimensions have spherical covariances and $0$ means. 
  Therefore, $d$ is identifiable
  as the smallest value such that the latter $D-d$ components all have spherical covariances and $0$ means. 
Furthermore, by specifying that
all mixing coefficients
  $\pi^{(k)}>0$ and that if $\mu^{(k_1)}=\mu^{(k_2)}$ then $\Sigma^{(k_1)}\neq\Sigma^{(k_2)}$,
  we guarantee that $K$ is identifiable
  as the smallest value such that any equivalent mixture components have been merged --
  the ``index of the economical representation'' from \cite{JCP2001}.
}

We establish our probability model for the extended ASE of $G \sim \text{SBM}(n, B, \bm\pi)$. Let the extended ASE be $\hat{Z} \in \mathbb{R}^{n \times D}$, then our conjecture states, for any $i \in [n]$,
\begin{equation}
  \label{eq: model of extended ASE}
  \hat{Z}_i \sim f(\cdot ; \theta^*(d_0, K_0))
\end{equation}
approximately for sufficiently large $n$, where $f(\cdot; \theta(d,K))$ is the density function defined in Model \ref{mod: GMM for undirected graphs}, $d_0$ is the true dimension of latent positions, $K_0$ is the true number of blocks, and $\theta^*(d, K)$ is the true underlying collection of parameters of the GMM. This conjecture states that the rows of the extended ASE are identically distributed as in the GMM.
In fact, it has been shown that the rows of ASE are not independent \citep{athreya2016limit, tang2018limit}. 
(See \cite{TangAsyEff} for one recent treatment of the dependency.)
However, for ease of analysis we will proceed in the consistency theorem and in the calculation of BIC by ignoring dependency, 
because the independence assumption makes the estimation tractable and simulation results demonstrate acceptable performance.

The GMMs from ASE are also {\it curved} families
\citep{athreya2016limit, tang2018limit};
we ignore this complexity here,
but see
\cite{ZPcurved} for an estimation algorithm which takes curvature into account (but still ignores dependence).

\section{Simultaneous Model Selection}
\label{sec: Simultaneous Model Selection}
\subsection{Simultaneous model selection framework}
The idea of simultaneous model selection is inspired by the model comparison presented in \cite{raftery2006variable}.
Assume $M_1$ and $M_2$ are models that both describe the same random vector. By Bayes' theorem, the posterior probability of the model is proportional to the product of the prior and the integrated likelihood, i.e. for $t = 1, 2$
\begin{equation}
    P(M_t|X) \propto P(M_t) P(X|M_t)
\end{equation}
where we call $P(X|M_t)$ the \textit{integrated likelihood} because it can be obtained by integrating over all the unknown parameters in the model, i.e.
\begin{equation}
    P(X|M_t) = \int P(X|\theta_t, M_t) P(\theta_t|M_t) d\theta_t.
    \label{eq: integrated likelihood 2}
\end{equation}
Since usually we assume no preference between the models, we can ignore the prior probability term $P(M_t)$ and just compare the integrated likelihoods. However, computing the integrated likelihood is impractical. Thus, we use the \textit{Bayesian information criterion} (BIC).

Now we consider $d$ and $K$ as the model parameters in the vertex clustering problem. Let $f(\cdot ; \theta(d, K))$ be the probability density function of the model which characterizes the distribution of $\hat{Z}_i$, the rows of extended ASE. We assume two models differ from each other if and only if they have distinct model parameters -- our Identifiability Assumption. (Gaussian mixture models have well-known non-identifiabilities---in particular, labeling of components is arbitrary---which are of no practical concern in most GMM inference tasks and do not concern us here.) So selecting a model from the family is equivalent to determining the model parameters. Now we can recast the model selection problem in the simultaneous model selection framework as follows. Provided we have a family of $D$-dimensional distributional models, each with a distinct pair of model parameters $(d,K)$ that determine the structure of the model, the model selection problem is to choose a model from amongst all $(d,K)$ pairs by comparing the values $\text{BIC}(\hat{Z}; d,K)$ evaluated on the observed $\hat{Z}$.

In the framework of \textit{simultaneous model selection} (SMS), a probability model $f(\cdot ; \theta(d, K))$ for the rows of the extended ASE is needed. The model parameter $d$ should play a similar role to the embedding dimension, which separates the informative dimensions and redundant dimensions in the extended ASE. 
The model parameter $K$ should be the number of mixture components in the mixture model. If we have such a family of models that well approximates the distribution of the extended ASE with an appropriate $(d,K)$, we can apply our simultaneous model selection procedure. Fortunately, Model \ref{mod: GMM for undirected graphs} exactly satisfies these requirements. To see this, let $G \sim \text{SBM}(n, B, \bm\pi)$ be the random graph and $\hat{Z} \in \mathbb{R}^{n \times D}$ be the corresponding extended ASE. Let $d_0 = \text{rank}(B)$ be the dimension of the latent position vectors, and let $K_0$ given by the dimension of $B$ be the number of blocks in the SBM. In Model \ref{mod: GMM for undirected graphs}, $d$ is the model parameter which determines the dimensionality of the informative part and $K$ is the model parameter which determines the number of components. Most importantly, the rows of $\hat{Z}$ approximately follow the distribution $f(\cdot ; \theta(d_0, K_0))$ by the existing theorem and our conjecture. Therefore if we use this family of models in simultaneous model selection, we expect maximizing BIC will provide good estimates of the model parameters $(d_0, K_0)$. In fact, if we assume that the rows of $\hat{Z}$ do asymptotically follow the distribution in the model, we can prove the consistency of the model parameter estimates obtained via our SMS procedure.

\subsection{Consistency of model parameter estimates}
We first define some notation. Let
\begin{equation}
  \label{eq: our model again}
  f(\cdot ; \theta(d, K)) = \sum_{k=1}^{K} \pi^{(k)} \varphi(\cdot; \mu^{(k)}, \Sigma^{(k)})
\end{equation}
be a family of GMM density functions for a $D$ dimensional random vector, as defined in Model \ref{mod: GMM for undirected graphs}, where $(d,K)$ are the model parameters which determine a specific class of densities. For given constants $d_0$ and $K_0$, let $\theta^*(d_0, K_0)$ be a set of given parameters in the density function (\ref{eq: our model again}). We define
\begin{equation}
    \theta^*(d, K) = \arg\min_{\theta(d, K) \in \Theta(d, K)} D_{\text{KL}}[f(\cdot ; \theta^*(d_0, K_0)) || f(\cdot ; \theta(d, K))]
\end{equation}
for all $d, K$. Here, $D_{\text{KL}}[g || h]$ is the Kullback-Leibler divergence of density $h$ from density $g$, defined as
\begin{equation}
    D_{\text{KL}}[g || h] = \mathbb{E}_{g(\cdot)}\left[ \log \left(\frac{g(X)}{h(X)}\right) \right] = \int \log \left(\frac{g(x)}{h(x)}\right)g(x) \text{d} x.
    \label{eq: KL divergence}
\end{equation}
Notice that this definition is self-consistent on $\theta^*(d_0, K_0)$, because $D_{\text{KL}}[g || h] \ge 0$ and equality holds if and only if $g = h$ almost everywhere, by the properties of KL divergence. We recall our Identifiability Assumption and say the model (\ref{eq: our model again}) is \textit{identifiable}
on the density $f(\cdot ; \theta^*(d_0, K_0))$, if for all $(d, K) \ne (d_0, K_0)$, $f(\cdot ; \theta^*(d, K)) \ne f(\cdot ; \theta^*(d_0, K_0))$. In other words, there are no identical density functions from the model with different $(d, K)$. (Again: GMMs do have some well-known and unconcerning non-identifiabilities; we are concerned with model complexity identifiability.) Let $\text{BIC} (\hat{Z}; d, K)$ denote the BIC evaluated on $\hat{Z}$ with model $f(\cdot ; \theta(d, K))$, i.e.
\begin{equation}
    \text{BIC} (\hat{Z}; d, K) = 2\sum_{i=1}^{n} \log[f(\hat{Z}_i; \hat\theta(\hat{Z}; d, K))] - \eta(d, K) \log(n)
    \label{eq: bic in theorem}
\end{equation}
where $\eta(d, K)$ is the number of parameters in the model, $n$ is the number of rows in $\hat{Z}$, and $\hat{\theta}(\hat{Z}; d, K)$ is the \textit{maximum likelihood estimator} (MLE) of the parameters from optimizing the loglikelihood
\begin{equation}
    \hat{\theta}(\hat{Z}; d, K) = \arg \max_{\theta(d, K) \in \Theta(d, K)} \frac{1}{n} \sum_{i = 1}^{n} \log f(\hat{Z}_i; \theta(d, K))
    \label{eq: MLE in theorem}
\end{equation}
where $\Theta(d, K)$ is the parameter space of the model with given $(d, K)$.

Using the notation defined above, we here state our theoretical result as follows.

\begin{theorem}[Consistency of model parameter estimates]
    Let $\{\hat{Z}^{(n)}\}_{n=1}^{\infty}$ be a sequence of random matrices, where each element $\hat{Z}^{(n)} \in \mathbb{R}^{n \times D}$ is a matrix with $n$ rows of $D$-dimensional random vectors. If
    
    a) the rows of $\hat{Z}^{(n)}$ are independently identically distributed according to (\ref{eq: our model again}), with parameter $\theta^*(d_0, K_0)$, i.e., for an arbitrary $n$,
    \begin{equation}
        \hat{Z}^{(n)}_i \sim f(\cdot ; \theta^*(d_0, K_0))
    \end{equation}
    i.i.d.\ for all $i \in [n]$;
    
    b) the model $f(\cdot ; \theta(d, K))$ satisfies our Identifiability Assumption on density $f(\cdot ; \theta^*(d_0, K_0))$; 
    
    c) for all $(d, K)$, the parameter space $\Theta(d, K)$ is a compact metric space; \\
    then the estimates of model parameters given by
    \begin{equation}
        (\hat{d}^{(n)}, \hat{K}^{(n)}) = \arg\max_{d \in [D], K \in [K_{\text{max}}]} \text{BIC} (\hat{Z}^{(n)}; d, K)
        \label{eq: definition of dhat khat}
    \end{equation}
    (with a constant $K_{\text{max}} \ge K_0$) will converge to the truth, i.e.
    \begin{equation}
        (\hat{d}^{(n)}, \hat{K}^{(n)}) \overset{p}{\longrightarrow} (d_0, K_0)
    \end{equation}
    as $n \to \infty$.
    \label{thm: main theorem}
\end{theorem}
We leave the proof of the theorem for the appendix. This theoretical support together with the practical advantages of simultaneous model selection motivate us to conduct vertex clustering via simultaneous model selection. 

\subsection{Algorithms for simultaneous model selection}
We present a model-based clustering algorithm via simultaneous model selection (SMS) with the Gaussian mixture model. 
The entire procedure consists of three phases. First, the ``parameter fitting'' phase. We compute the MLE -- the maximum likelihood estimator -- in the GMM for each pair $(d,K)$. The MLEs are used to complete the density function while evaluating the likelihood on the data. Second, the ``model selection'' phase. We compute the BIC values for all $(d, K)$ pairs, then choose the one with the largest BIC as the model parameter estimate given the data. Finally, the ``clustering'' phase. The likelihoods of all the data points are evaluated on the selected model with fitted parameters. Labels are assigned to each point by the \textit{maximize a posterior} (MAP) rule. A summary of the SMS algorithm is provided in the Algorithm \ref{alg: oh MCG} display.

\begin{algorithm}[ht]
    \caption{SMS -- Simultaneous Model Selection}
    \label{alg: oh MCG}
    \textbf{Input:} The adjacency matrix $A \in \mathbb{R}^{n \times n}$; an upper bound $D$ on embedding dimension; an upper bound $K_{\text{max}}$ on mixture complexity
    \begin{algorithmic}[1]
        \Function{SMS}{$A, D, K_{\text{max}}$}
            \State Apply extended ASE on $A$ with dimension $D$: $\hat{Z} \gets \hat{U}_{[D]} \hat{\Lambda}_{[D]}^{\frac{1}{2}}$
            \Loop
                \State Compute MLEs $\hat{\theta}(\hat{Z};d, K) = \{\hat{\pi}^{(k)}, \hat{\mathbf{\mu}}^{(k)}, \hat{\Sigma}^{(k)}\}_{k = 1}^{K}$ for Model \ref{mod: GMM for undirected graphs}
            \EndLoop
            \State $(\hat{d}, \hat{K}) \gets \arg\max_{d \in [D] ,K \in [K_{\text{max}}]} \text{BIC}(\hat{Z}; d,K)$
            \State $\hat{\tau}_i = \arg \max_k \left\{\frac{\hat{\pi}^{(k)} \varphi(\hat{Z}_i;\hat{\mu}^{(k)}, \hat{\Sigma}^{(k)})}{\sum_{k'=1}^{\hat{K}} \hat{\pi}^{(k')} \varphi(\hat{Z}_i;\hat{\mu}^{(k')}, \hat{\Sigma}^{(k')})}\right\}$
        \EndFunction
    \end{algorithmic}
    \textbf{Output:} The clustering labels $(\hat{\tau}_1, \dots, \hat{\tau}_n)$
\end{algorithm}

Although we believe that simultaneous model selection has advantages compared to its sequential ounterpart, it is unclear whether including the redundant dimensions of the extended ASE in the clustering phase is preferable. The reasoning can be explained by two aspects. First, the redundant dimensions may contain little information for the clustering---as indicated by the model, and by the
simulations in the preceding section, only a single variance term contains potential
clustering information. Second, choosing a smaller dimension in the clustering task may lead to better performance, especially for a small number of observations, due to the bias-variance tradeoff \citep{jain2000statistical}. 
This motivates a variation in the third phase of the SMS algorithm. To be specific, in phase 1 and phase 2, Model \ref{mod: GMM for undirected graphs} and the extended ASE $\hat{Z}$ are utilized just to find the estimate of embedding dimension. In phase 3, we can now truncate the extended ASE to the dimension $\hat{d}$ which is estimated by SMS. In this alternative context, redundant dimensions do not take part in the clustering procedure. Thus we may apply the traditional model-based clustering algorithm with regular GMM on the truncated embedding $\hat{Z}_{\hat{d}}$. Notice that the embedding dimension is determined by the SMS procedure, so the clustering results could be remarkably different than the algorithm under a sequential model selection framework. We call this algorithm SMS-Reduced, inspired by model-based clustering by GMM with reduced embedding dimension determined via SMS. The outline of the steps of SMS-Reduced is provided in the Algorithm \ref{alg: MCEG} display.

\begin{algorithm}[ht]
    \caption{SMS-Reduced}
    \label{alg: MCEG}
    \textbf{Input:} The adjacency matrix $A \in \mathbb{R}^{n \times n}$; an upper bound $D$ on embedding dimension; an upper bound $K_{\text{max}}$ on mixture complexity
    \begin{algorithmic}[1]
        \Function{SMS-Reduced}{$A, D, K_{\text{max}}$}
            \State Apply extended ASE on $A$ with dimension $D$: $\hat{Z} \gets \hat{U}_{[D]} \hat{\Lambda}_{[D]}^{\frac{1}{2}}$
            \Loop
                \State Compute MLEs $\hat{\theta}(\hat{Z};d, K) = \{\hat{\pi}^{(k)}, \hat{\mathbf{\mu}}^{(k)}, \hat{\Sigma}^{(k)}\}_{k = 1}^{K}$ for model \ref{mod: GMM for undirected graphs}
            \EndLoop
            \State $\hat{d} \gets \arg\max_{d \in [D] ,K \in [K_{\text{max}}]} \text{BIC}(\hat{Z}; d,K)$
            \State Truncate the ASE: $\hat{Z}_{[\hat{d}]} \gets \hat{U}_{[\hat{d}]} \hat{\Lambda}_{[\hat{d}]}^{\frac{1}{2}}$
            \State $(\hat{\tau}_1, \dots, \hat{\tau}_n) \gets \text{GMM} \circ \text{BIC} (\hat{Z}_{[\hat{d}]})$
        \EndFunction
    \end{algorithmic}
    \textbf{Output:} The clustering labels $(\hat{\tau}_1, \dots, \hat{\tau}_n)$
\end{algorithm}


\begin{algorithm}[ht]
    \caption{SMS Two-Step}
    \label{alg: twostep}
    \textbf{Input:} The adjacency matrix $A \in \mathbb{R}^{n \times n}$; an upper bound $D$ on embedding dimension; an upper bound $K_{\text{max}}$ on mixture complexity
    \begin{algorithmic}[1]
        \Function{TwoStep}{$A, D, K_{\text{max}}$}
            \State Apply extended ASE on $A$ with dimension $D$: $\hat{Z} \gets \hat{U}_{[D]} \hat{\Lambda}_{[D]}^{\frac{1}{2}}$
            \Loop
                \State Step 1: Compute GMM on $\hat{Z}_d$ using model based clustering to choose $\hat{K}$  \citep{fraley2002model}
                \State Step 2: Use the fitted mixture to estimate the $\sigma^2_j$
                of the redundant part. Compute BIC for the resulting model
            \EndLoop
            \State $\hat{d} \gets \arg\max_{d \in [D] ,K \in [K_{\text{max}}]} \text{BIC}(\hat{Z}; d,K)$
            \State $(\hat{\tau}_1, \dots, \hat{\tau}_n) \gets \text{GMM} \circ \text{BIC} (\hat{Z}_{[\hat{d}]})$
        \EndFunction
    \end{algorithmic}
    \textbf{Output:} The clustering labels $(\hat{\tau}_1, \dots, \hat{\tau}_n)$
\end{algorithm}

There is another implementation that we consider here, which we refer to as the ``two-step''
algorithm. It is nearly equivalent to the reduced algorithm---and in fact is identical to it
in those cases where the two models produce the same $\hat{d},\hat{K}$.
The loop in Algorithm \ref{alg: twostep} is over $d$ alone. There is indeed a loop over $K$
in the model based clustering, which halts when the BIC value decreases.
Note that this approach does not
necessarily produce the maximum likelihood solution; however, in the event the estimates of
$d$ and $K$ are the same as those of the reduced model, the resulting models are identical.
This eliminates the extra step of fitting the GMM to the $\hat{d}$ dimensional embedding,
for a computational advantage. This approach makes explicit that the informative
part of the model need not be the full, unconstrained model described above; by using
``mclust'' explicitly in step 1, we have the ability to use the full range of models 
implemented in the \texttt{mclust} R package \citep{mclust}. In this paper we consider only the full model described above
and do not investigate modifications that would allow for a wider range of constrained models,
but it would be simple to implement these within the two-step procedure.

\section{Experimental Results}
\label{sec: Experimental Results}
We evaluate the performance of the simultaneous model selection algorithms via simulation experiments, and illustrate our methods via application to a collection of brain graphs.


We compare our methods with the sequential BIC $\circ$ ZG method,
first choosing the dimension a la \cite{zhu2006automatic}
and then using BIC in the GMM to choose the number of clusters.
Deciding which scree plot elbow to use is always subjective in practice, so we will consider ZG1, ZG2 and ZG3, the ZG algorithm which takes the 1st, 2nd and 3rd elbows respectively.
For $\ell = 1,2,3$
we use the notation
ZG$\ell$;
hence BIC $\circ$ ZG$\ell$ provides three sequential model selection competitors.
Notice that even if one ZG$\ell$ (or corresponding BIC $\circ$ ZG$\ell$) method outperforms our proposed simultaneous methods in a specific setting, this does not mean that the ZG algorithm is superior to ours because the optimal $\ell$ will be different in a different setting. We will see this in the simulations. 
We apply the \texttt{mclust} R package \citep{mclust} to perform the BIC selection, always using the ``full covariance'' model for consistency throughout. Additionally, we also consider three well-known heuristic vertex clustering methods for comparison: the Louvain algorithm proposed in \cite{blondel2008fast}; the Walktrap algorithm proposed in \cite{pons2005computing}; and the Infinite Relational Model \citep{irm}.

There are numerous criteria to evaluate the performance of a clustering result, including Jaccard \citep{jaccard1912distribution}, Rand index \citep{hubert1985comparing}, normalized mutual information \citep{danon2005comparing} and variation of information \citep{meilua2007comparing}. Of these, we choose the well known \textit{adjusted Rand index} (ARI) as the measure of the similarity between the clustering result and the ground truth labels---in the simulations the ground truth
is known, and it is this that is used to compute the ARI. As a corrected-for-chance version of the Rand index, ARI normalizes the Rand index so that the expected value between a random cluster and the ground truth is zero. The maximum value of ARI is $1$, which indicates perfect agreement of two partitions. So a larger ARI means the clustering is performing better.

\subsection{Numerical results on synthetic data}
\begin{figure}[ht]
    \centering
    \includegraphics[width=0.55\columnwidth,clip=true]{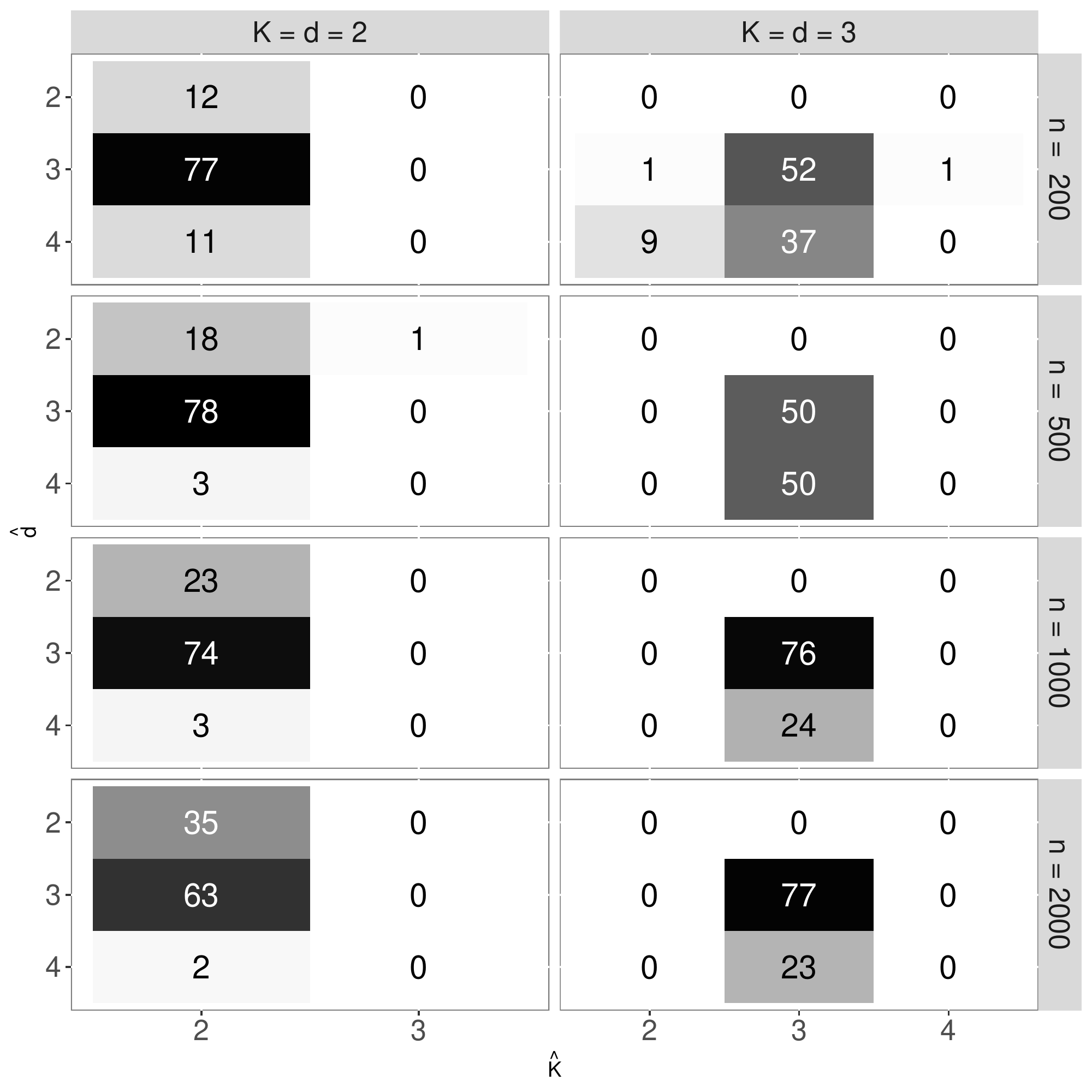}
\caption{
Simultaneous Model Selection (SMS) simulation results from $100$ Monte Carlo trials with the 2- and 3-block models described in Section \ref{sec: Distributional results for extended ASE}.}
    \label{tab:bsims}
\end{figure}

 First, we consider Monte Carlo simulation for the two- and three-block models
described in Section \ref{sec: Distributional results for extended ASE}. 
Recall that both models have a full-rank $B$ matrix, so
$K=d=2$ for the former and
$K=d=3$ for the latter.
We used $(D,K_{max})=(6,6)$ in this simulation, with $100$ Monte Carlo replicates.
 The results are depicted in Figure \ref{tab:bsims}. There is a tendency for the SMS method to choose $\hat{d}$ slightly
larger than the rank of $B$. 
(This is an acceptable bias: {\it under}estimating $d$ can have catastrophic consequences for subsequent inference, while {\it over}estimating $d$ has a relatively minor impact.
However,
the correct $K$ is chosen nearly always (all but once for the 2-block model, and all but
$11$ times for the 3-block model, out of a total of $400$ simulations for each model).

Next we generate a graph $G$ from a stochastic block model SBM($n, B, \bm\pi$) by specifying the block probability matrix $B$, prior block probability $\bm\pi$, and number of vertices $n$. The adjacency matrix $A \in \mathbb{R}^{n \times n}$ represents $G$. Then we apply the extended adjacency spectral embedding on the graph, denoted by $\hat{Z} \in \mathbb{R}^{n \times D}$. For simplicity, we fix $D = 8$. Let $n=500$, $$B =
\begin{bmatrix}
  0.2  &p \\
  p    &0.1
\end{bmatrix}$$ and $\bm\pi = (0.5, 0.5)$. We vary $p$ to change the angle between two latent vectors.

\begin{figure}[ht]
    \centering
    \includegraphics[width=0.98\columnwidth,clip=true]{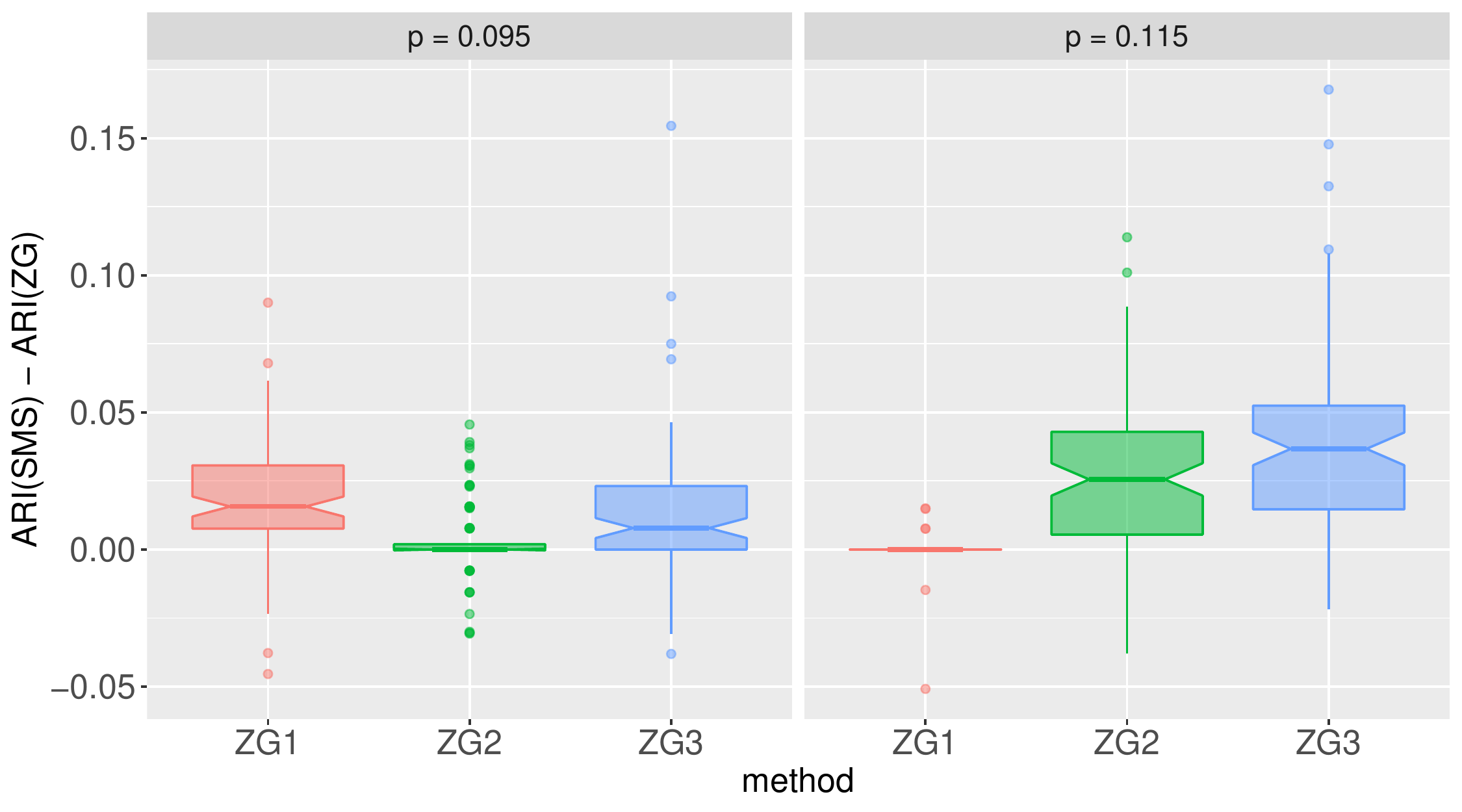}
    \caption[The difference of adjusted Rand index (ARI) between our simultaneous model selection algorithm (SMS-reduced) and BIC $\circ$ ZG methods in full rank case]{The difference of adjusted Rand index (ARI) between our simultaneous model selection algorithm (SMS-reduced) and BIC $\circ$ ZG methods shown in boxplots. Random graphs are generated from a 2-block SBM with a full rank block probability matrix. The number of vertices is fixed as $n=500$. The between block probability $p$ varies: (a) $p=0.095$, (b) $p=0.115$.}
    \label{fig: ch4_difference of ari}
\end{figure}

\begin{figure}[p]
    \centering
    \includegraphics[width=0.9\columnwidth,clip=true]{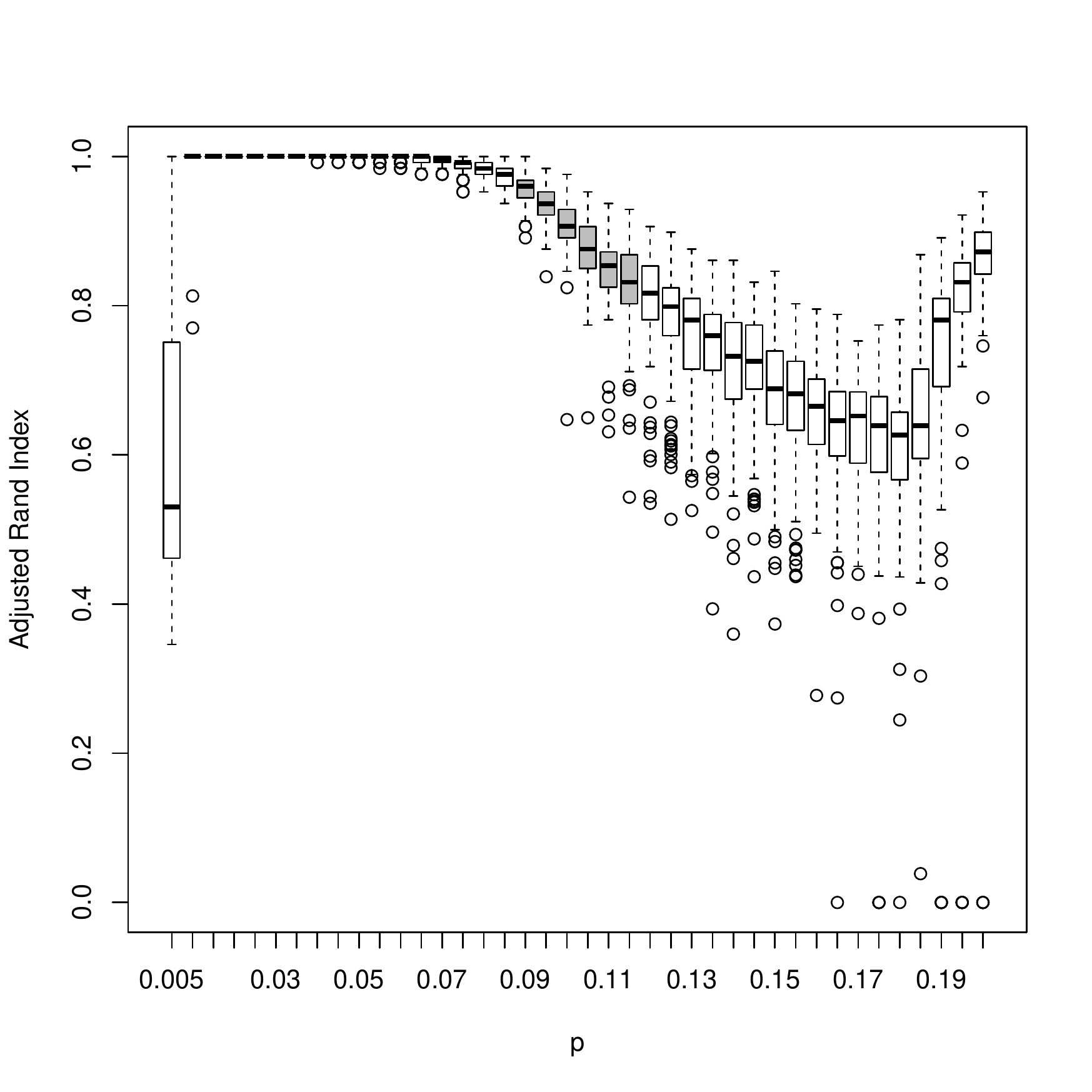}
    \caption[Boxplots of ARI of 100 Monte Carlo trials for the reduced method.]{Boxplots of adjusted Rand index (ARI) of 100 Monte Carlo trials for the simultaneous model selection (SMS) method. The random graph with $n = 500$ vertices is generated from a 2-block SBM with block probability matrix [0.2, $p$; $p$, 0.1]. 
    The parameter $p$ is varied from $0.005$ to $0.2$. The gray boxes correspond to the
    region of the $p$ parameter used in the subsequent experiment described in Figure \ref{fig: all methods}.}
    \label{fig: reduced}
\end{figure}

For this experiment we consider the SMS-reduced algorithm, which we refer to as ``our algorithm''.
Figure \ref{fig: ch4_difference of ari} shows the difference of ARI (computed using the
ground truth provided by the simulations) between 
our algorithm and the BIC $\circ$ ZG methods in boxplots -- $100$ Monte Carlo trials each produces paired results allowing for analysis of differences. A value larger than $0$ means our algorithm has a higher ARI than the corresponding BIC $\circ$ ZG method in that Monte Carlo replicate. 
Figure \ref{fig: ch4_difference of ari} (left panel) shows the result under the setting with $p=0.095$. We find our algorithm outperforms BIC $\circ$ ZG1 and BIC $\circ$ ZG3, and performs nearly the same as BIC $\circ$ ZG2. We perform a sign test for the paired differences of ARI, where the null hypothesis is that the two methods are equally good or BIC $\circ$ ZG is better ($\theta \le 0.5$ with respect to Binomial distribution), and the alternative hypothesis is that our method is better ($\theta > 0.5$). 
The p-values for our algorithm comparing to BIC $\circ$ ZG1, BIC $\circ$ ZG2 and BIC $\circ$ ZG3 are $<10^{-6}$, $0.04$ and $<10^{-6}$ respectively. The small p-values suggest that our algorithm is superior, surely, to BIC $\circ$ ZG1 and BIC $\circ$ ZG3. 
In Figure \ref{fig: ch4_difference of ari} (right panel), we use $p=0.115$ in the $B$ matrix. In this case, the p-values of a sign test for 
our algorithm compared to BIC $\circ$ ZG1, BIC $\circ$ ZG2 and BIC $\circ$ ZG3 are $0.34$, $<10^{-6}$ and $<10^{-6}$ respectively. In this case, 
our algorithm has similar performance to BIC $\circ$ ZG1, but outperforms BIC $\circ$ ZG2 and BIC $\circ$ ZG3. In both cases, our algorithm has joint best performance with respect to ARI. In contrast, none of the BIC $\circ$ ZG methods win in both cases.
The analogous t-test $p$-values agree across the board.
Considering that in practice we need to fix an elbow in BIC $\circ$ ZG methods without knowing the ground truth, the SMS algorithm provides a robust solution.

\begin{figure}[ht]
    \centering
    \includegraphics[width=0.85\columnwidth,clip=true]{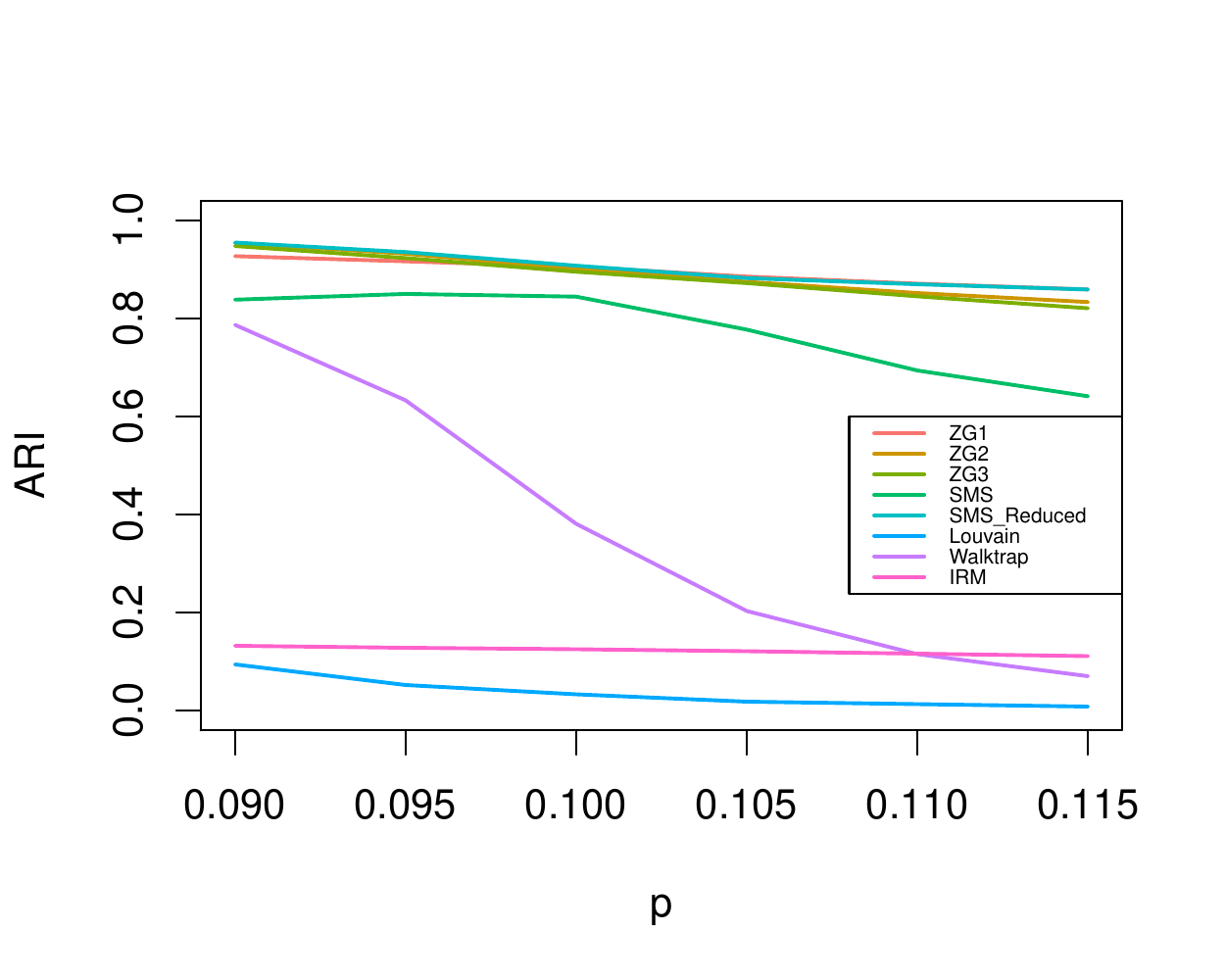}
    \caption[The mean of ARI of 100 Monte Carlo trials for different methods.]{The mean of adjusted Rand index (ARI) of 100 Monte Carlo trials for different methods. The random graph with $n = 500$ vertices is generated from a 2-block SBM with block probability matrix [0.2, $p$; $p$, 0.1]. The parameter $p$ is varying from $0.09$ to $0.115$.}
    \label{fig: all methods}
\end{figure}

\begin{figure}[ht]
    \centering
    \subfigure[mean of adjusted Rand index (ARI)]{
    \includegraphics[width=0.48\columnwidth,clip=true]{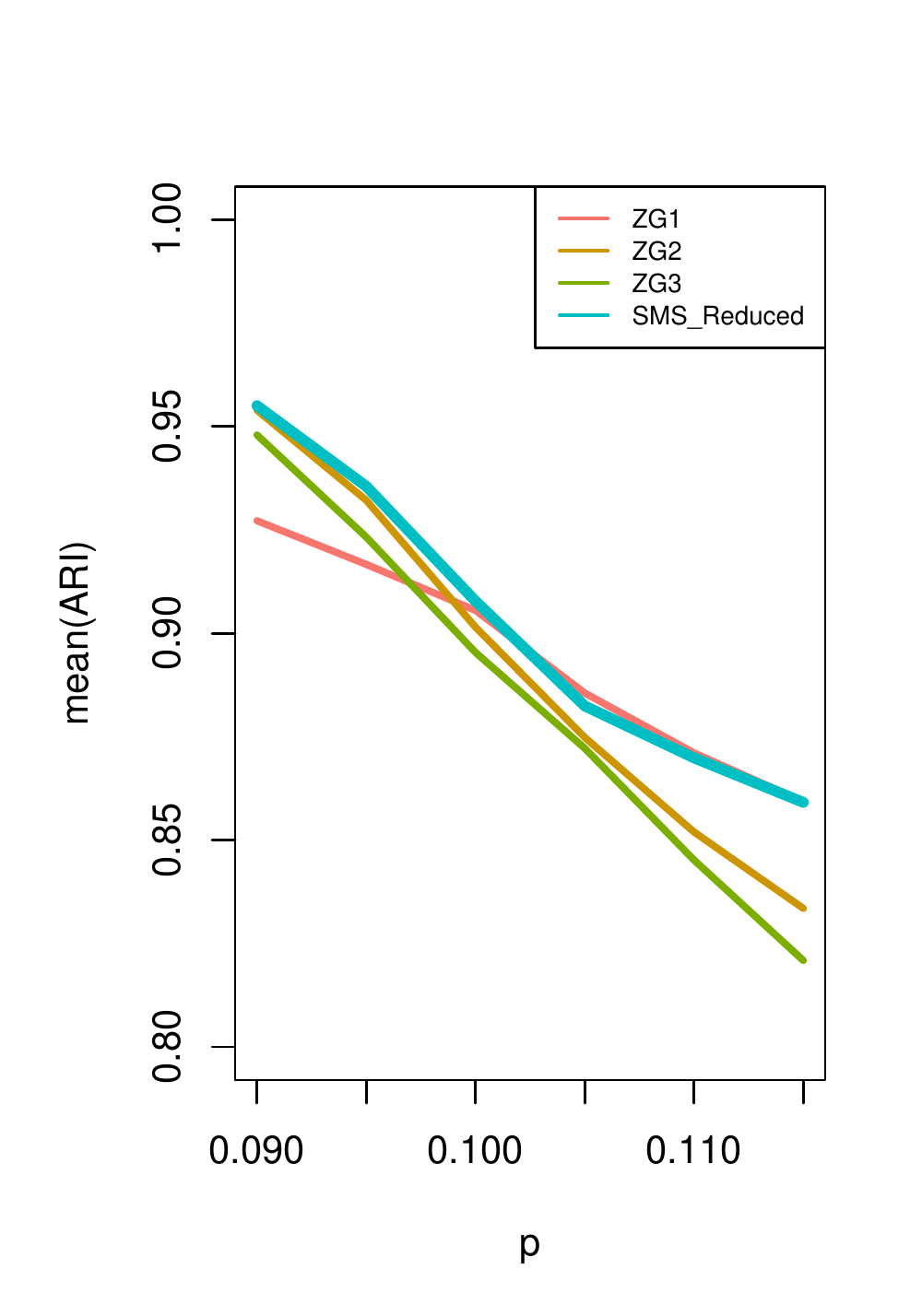}}
    \subfigure[mean of $\hat{d}-d$]{
    \includegraphics[width=0.48\columnwidth,clip=true]{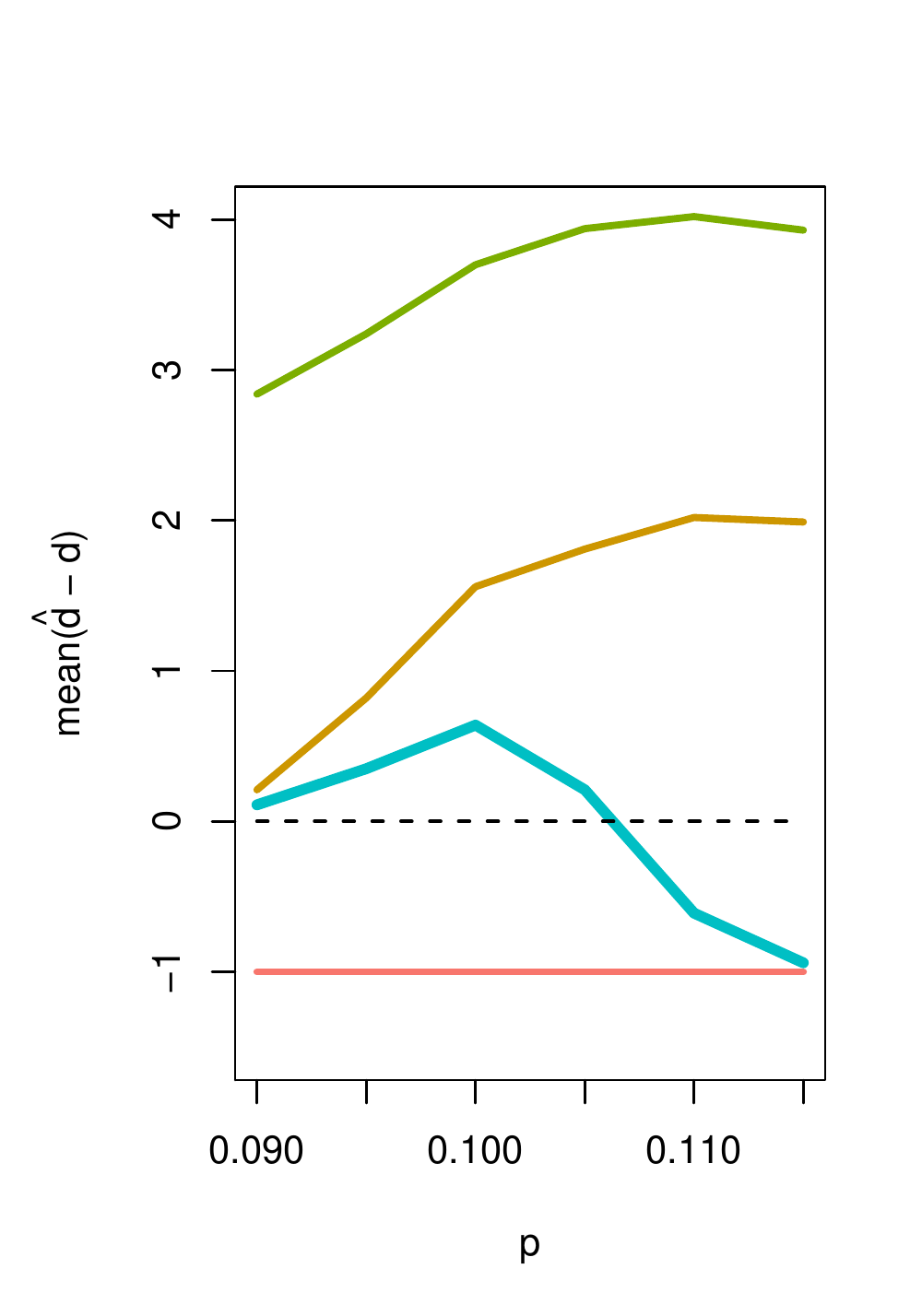}}
    \caption[The mean of ARI and $\hat{d}$ for varying p]{The mean of ARI and $\hat{d}$ for varying $p$: (a) mean of adjusted Rand index (ARI); (b) mean of $\hat{d} - d$. The random graph with $n = 500$ vertices is generated from a 2-block SBM with block probability matrix [0.2, $p$; $p$, 0.1]. The parameter $p$ is varying from $0.09$ to $0.115$.}
    \label{fig: ch4_difference of ari mceg}
\end{figure}

Figure \ref{fig: reduced} shows the distribution of adjusted Rand index (ARI) for the simultaneous model selection (SMS) algorithm for a range
of values $p$ in the $B$ matrix. The gray boxes in the figure indicate those for which more extensive experiments are 
reported, comparing the reduced and full algorithms to other methods. The reason for the first
low box with median near $50$\% 
is the fact that for very small $p$ there is a high probability of the graphs being disconnected,
in which case simple spectral embedding is inappropriate.

Figure \ref{fig: all methods} shows the mean of ARI for all methods, including the existing heuristic Louvain, Walktrap, and Infinite Relational Model (IRM) algorithms. The random graph with $n = 500$ vertices is generated from a 2-block SBM with block probability matrix [0.2, $p$; $p$, 0.1]. The parameter $p$ is varying from $0.09$ to $0.115$. We observe that the Louvain, Walktrap, and IRM algorithms do not perform well for large $p$, so we may conclude that these three heuristic vertex clustering algorithms are not suitable for specific SBM graphs. To have a detailed look, Figure \ref{fig: ch4_difference of ari mceg} shows the mean of ARI for SMS and the ZGs. 
For small $p$ ZG1 is the best of the ZGs, while for large $p$ ZG2 is the best of the ZGs;
in all cases, SMS is as good as the best ZG.
This demonstrates precisely the robustness sought.
In Figure \ref{fig: ch4_difference of ari mceg}(a), all methods have decreasing ARI as $p$ increases. This is because the angle between two latent vectors become smaller, so the cluster centers get closer. Out of all the methods, our algorithm performs well for all $p$'s. In Figure \ref{fig: ch4_difference of ari mceg}(b), the mean of $\hat{d}-d$ is plotted. We can see that the SMS algorithm is the closest one to zero, which means it estimates $\hat{d}$ better than the other methods.

\subsection{Demonstration on connectomes}
We demonstrate the performance of the simultaneous model selection procedure on a real data set of human connectomes.
A connectome represents the brain as a network consisting of neurons (or collections thereof) as vertices and synapses (or structural connections) as edges. It is fundamentally helpful to unlock the structural and functional unknowns in the human brain in cognitive neuroscience and neuropsychology by studying the topological properties of connectomes. 
For this demonstration, the raw data is collected by diffusion magnetic resonance imaging (dMRI), which can represent the structural connectivity within the brain. The macro-scale connectomes are estimated by NeuroData's MRI to graphs (NDMG) pipeline \citep{kiar2018high}, which is designed to produce robust and biologically plausible connectomes across studies, individules and scans. 
As the output of the NDMG pipeline, the brain graphs are generated. The vertices of the graph represent \textit{regions of interest} identified by spatial proximity, and the edges of the graph represent the connection between regions via tensor-based fiber streamlines. Specifically, there is an edge for a pair of regions if and only if there is a streamline passing between them. 
The graph is undirected since the raw dMRI data has no direction information. 
For more details,
we refer the readers to \cite{kiar2018high}.

This specific data set
consists of 114 connectomes (57 subjects with 2 scans each), with $1215$ vertices for each graph. There are two attributes for each vertex (region of interest): hemisphere, either left or right;
and tissue type, either gray or white.
(In fact, the attributes can also take value ``other'';
for ease of illustration, we consider only regions labeled left or right hemisphere and gray or white tissue, yielding an induced subgraph from the original connectome by deleting the vertices with label ``other'' in hemisphere or tissue attributes.)
Then we extract the largest connected component of that subgraph so as to facilitate the spectral embedding. This yields 114 connected undirected graphs, with approximately 760 vertices for each graph. As described, each vertex is assigned two labels, hemisphere and tissue type. These are treated as ground truth for the structure in the graphs. 
We note that these labels do {\it not} necessarily correspond to {\it cluster} structure of the vertices, and so it is difficult to claim that one method is better than another for any {\it true} clustering structure.
At most we can claim that clustering the brain regions without using anatomical information provides results in agreement with the anatomy.
This does not necessarily translate to any true clustering structure that might be at odds with anatomy.
This issue is inherent in clustering the nodes of these graphs, a la the ``Two Truths'' phenomenon described in \cite{priebe2018two}.

\begin{figure}[h!]
    \centering
    \includegraphics[width=0.9\columnwidth,clip=true]{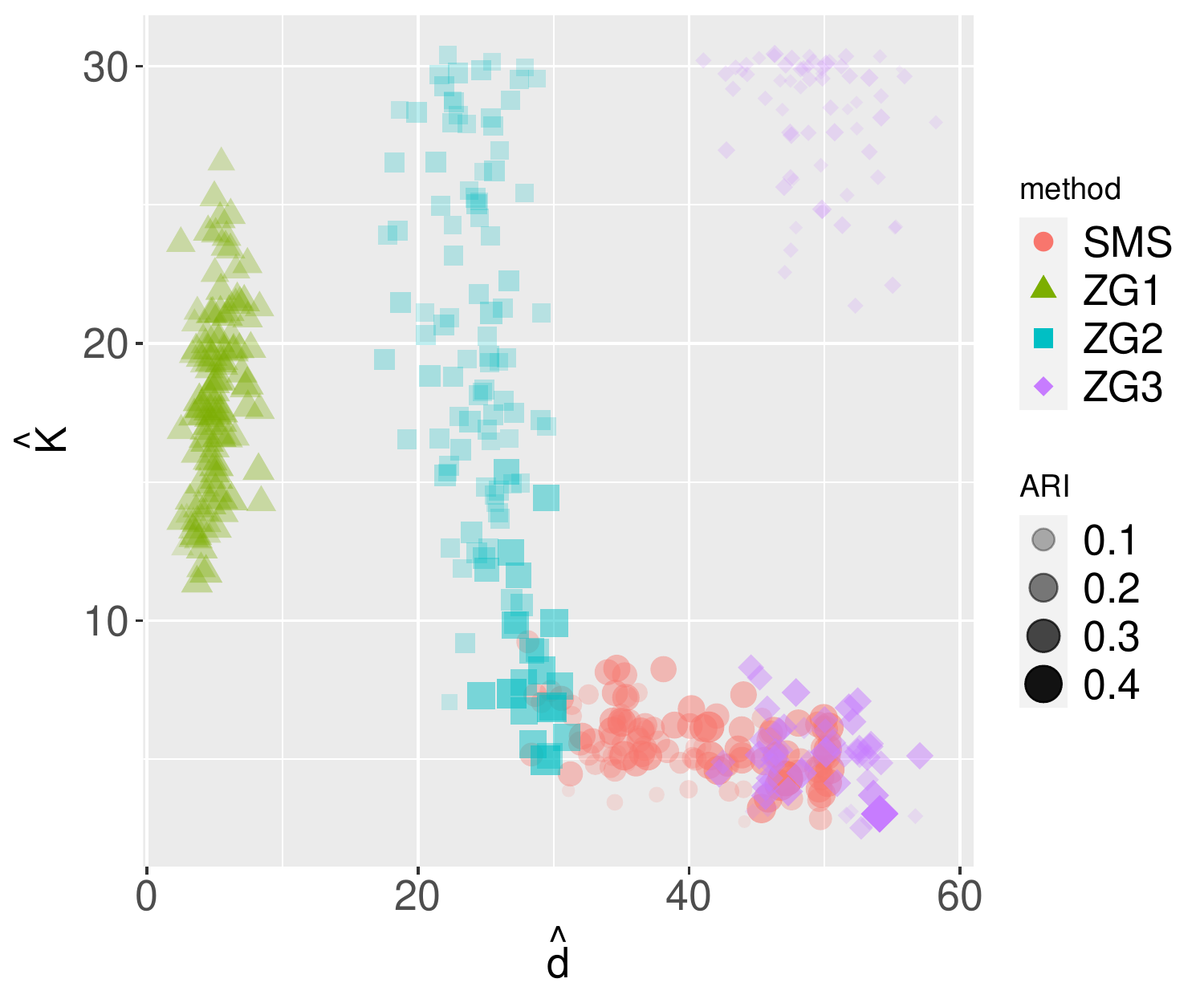}
    \caption[The estimates of the model parameter estimates $(\hat{d}, \hat{K})$ for 114 connectomes]{The estimates of the model parameter estimates $(\hat{d}, \hat{K})$ for 114 connectomes. $D = 100$ and $K_{\text{max}} = 30$. For each graph, four estimates by different methods are presented in different colors. The adjusted Rand index (ARI) values against
    (left-gray, left-white, right-gray or right-white) from combining the hemisphere and tissue
    are indicated by the sizes of the dots. (The coordinates of the points are slightly perturbed so as to view the occlusion.) 
    Simultaneous model selection (SMS) gives a smaller and more concentrated estimate of number of clusters 
    and the largest average ARI.}
    \label{fig: ch6_dhat khat}
\end{figure}

We apply clustering methods on the 114 graphs to compare our SMS algorithms (both full and reduced) with the sequential BIC $\circ$ ZG.
The ARI is calculated by comparing the clustering results with three separate versions of ground truth, namely hemisphere, tissue, and the combination of the two. Specifically, each vertex of a connectome is assigned a label left or right from the 2-cluster attribute hemisphere, and a label gray or white from the 2-cluster attribute tissue. We also assign a label (left-gray, left-white, right-gray or right-white) from the 4-cluster attribute by combining the hemisphere and tissue. 

\begin{figure}[h!]
    \centering
    \subfigure[SMS - BIC $\circ$ ZG1]{
    \includegraphics[width=0.30\columnwidth,clip=true]{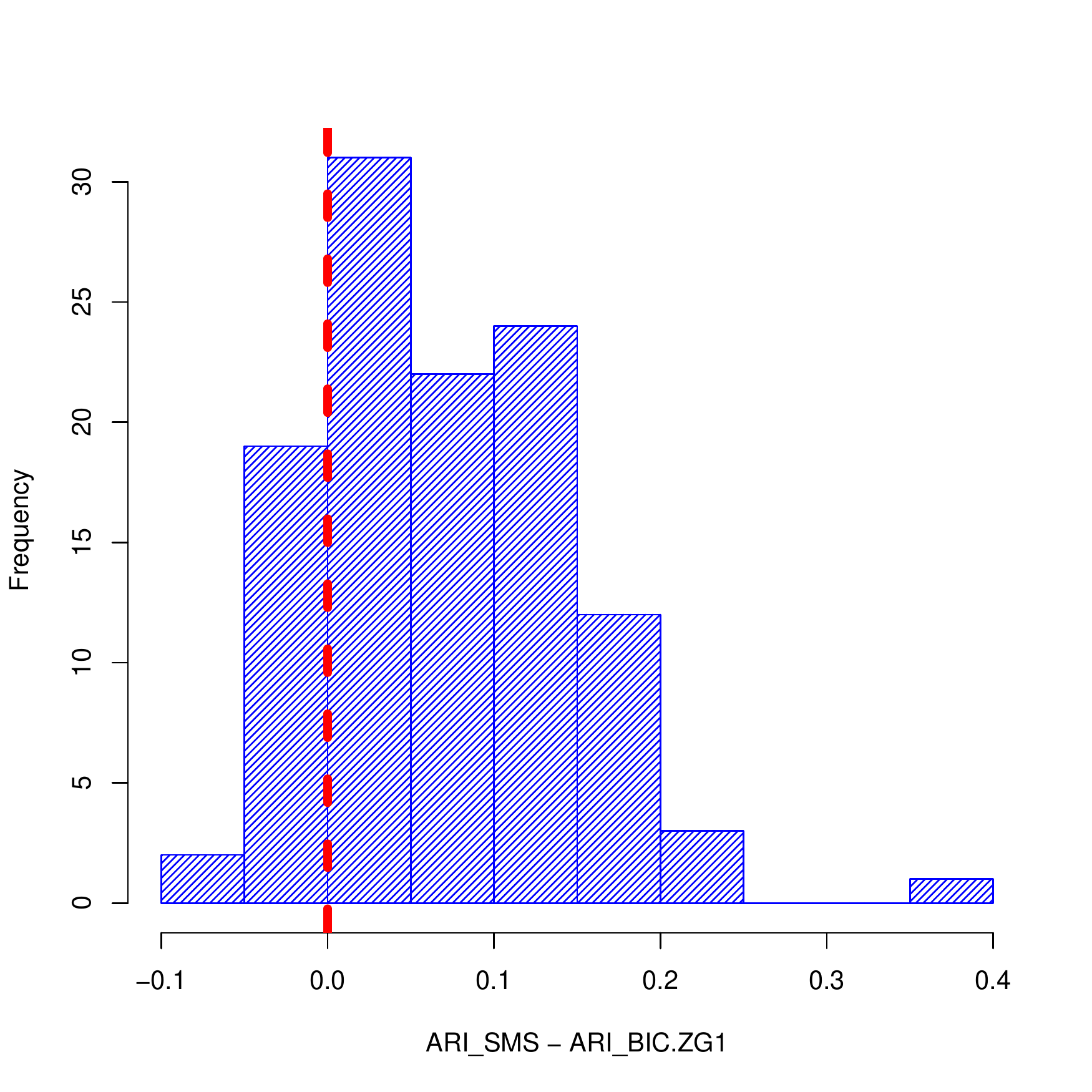}}
    \subfigure[SMS - BIC $\circ$ ZG2]{
    \includegraphics[width=0.30\columnwidth,clip=true]{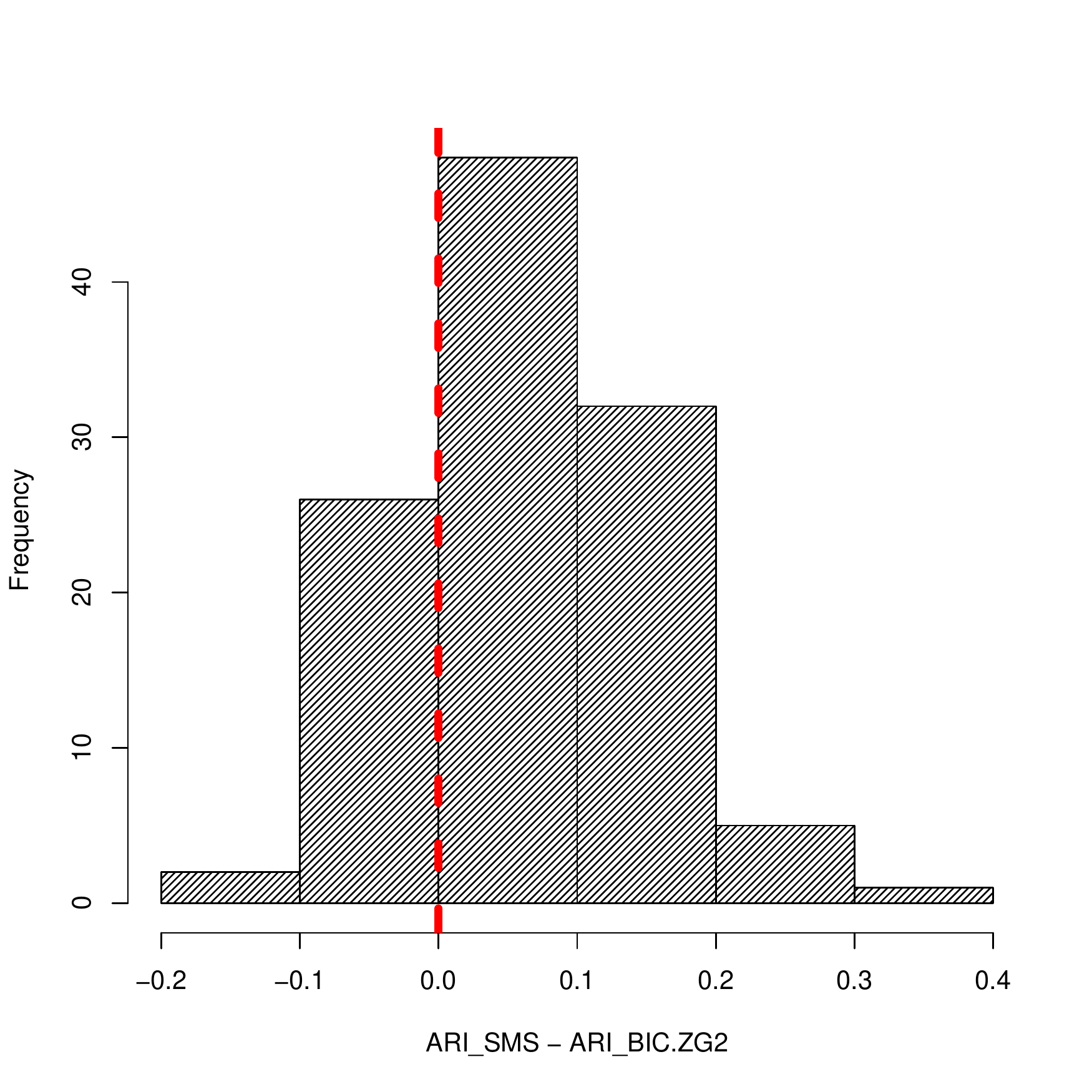}}
    \subfigure[SMS - BIC $\circ$ ZG3]{
    \includegraphics[width=0.30\columnwidth,clip=true]{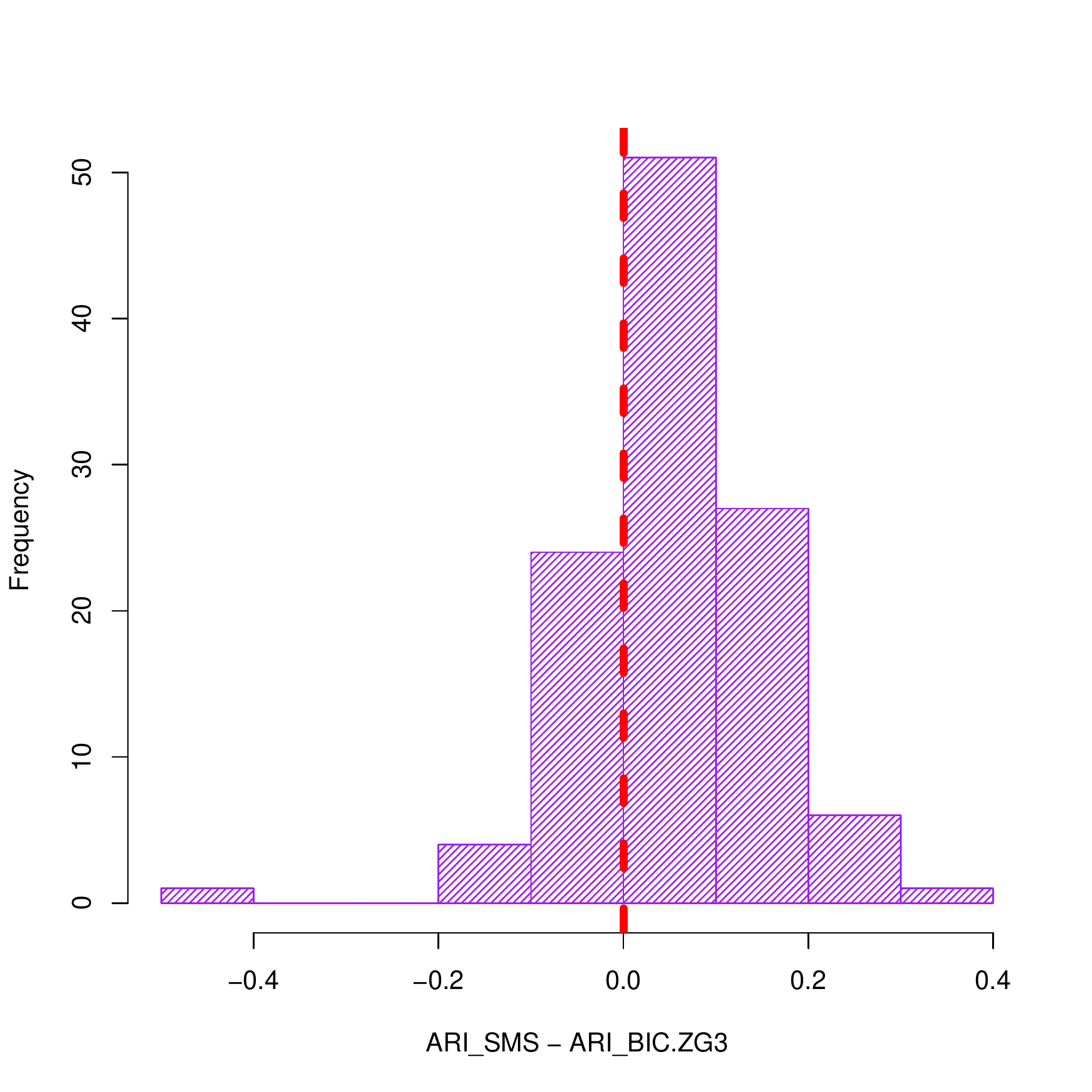}}
    \subfigure[SMS-R - BIC $\circ$ ZG1]{
    \includegraphics[width=0.30\columnwidth,clip=true]{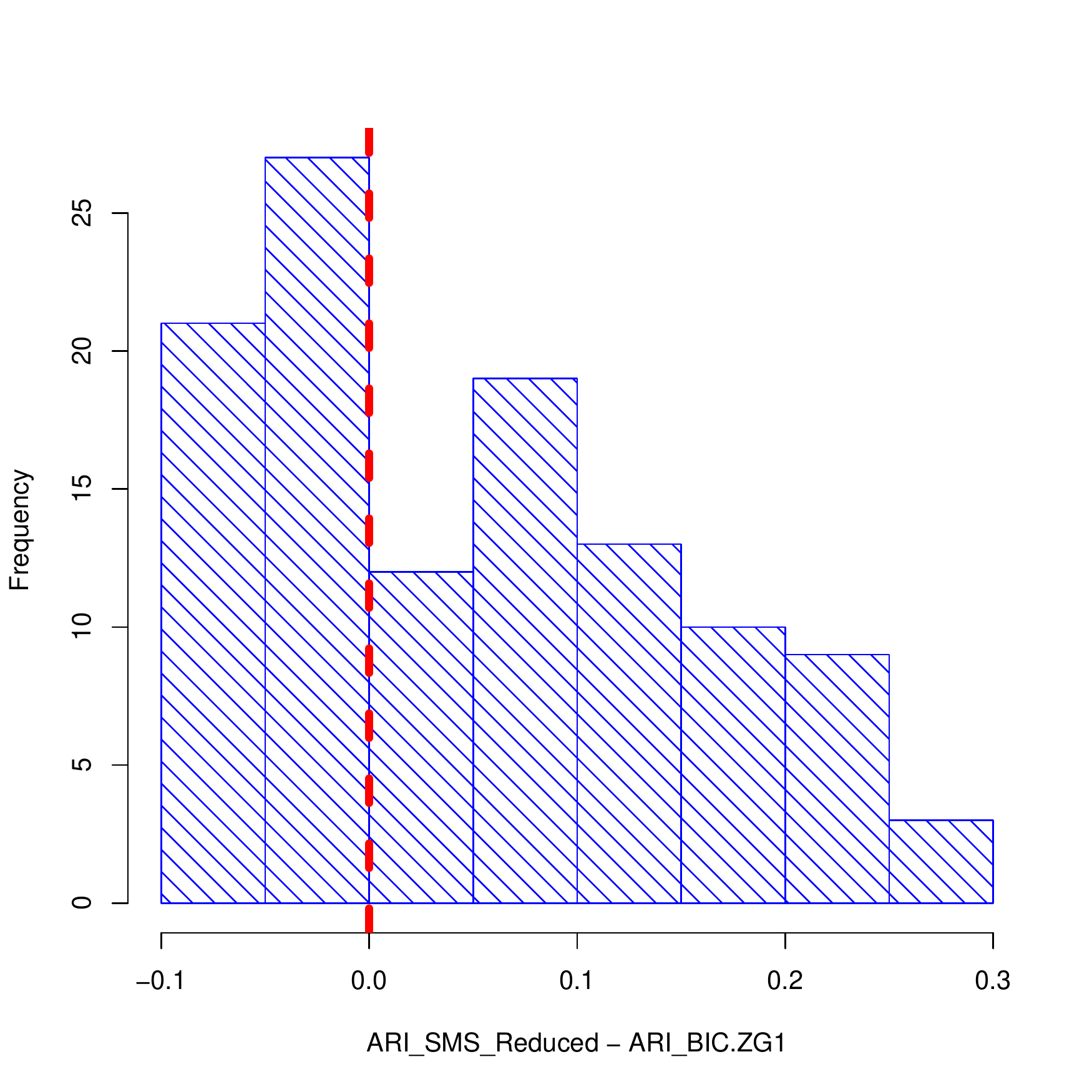}}
    \subfigure[SMS-R - BIC $\circ$ ZG2]{
    \includegraphics[width=0.30\columnwidth,clip=true]{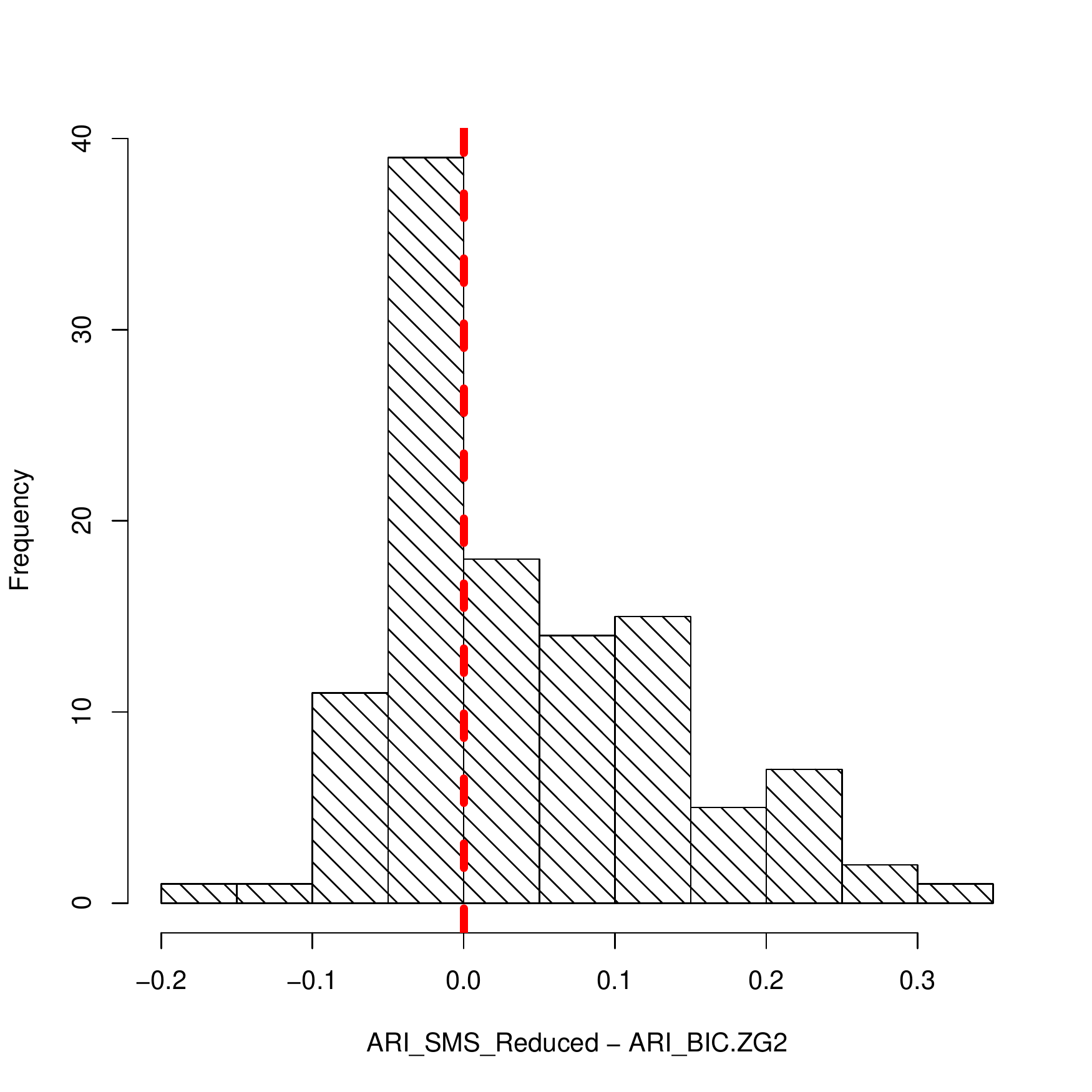}}
    \subfigure[SMS-R - BIC $\circ$ ZG3]{
    \includegraphics[width=0.30\columnwidth,clip=true]{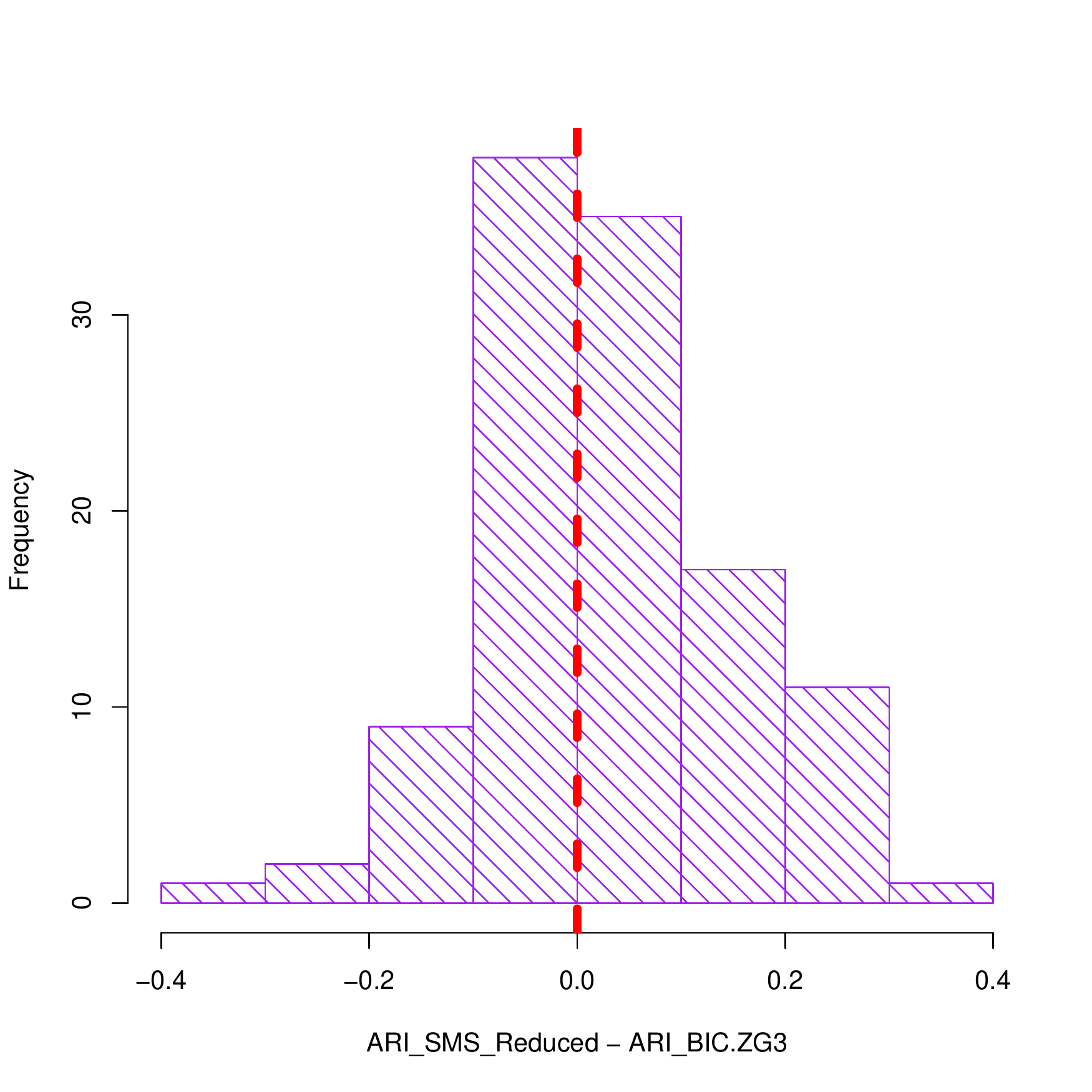}}
    \caption[Illustration of the difference of ARI between MCG/MCEG and BIC $\circ$ ZG methods]{Illustration of the difference of adjusted Rand index (ARI) between simultaneous model selection (SMS) and the sequential BIC $\circ$ ZG. The ARIs are computed against the ground truth in attribute hemisphere. The differences are taken pairwise for all 114 graphs. (a)-(c) show the histogram of the 114 differences between SMS and BIC $\circ$ ZGs, while (d)-(f) show the histogram of those between SMS-Reduced and BIC $\circ$ ZGs. More positive values in the histogram indicates stronger evidence that SMS outperforms BIC $\circ$ ZG. While SMS dominates in all cases, SMS-Reduced wins slightly with higher average ARI.}
    \label{fig: ch6_hist difference of ARIs}
\end{figure}

Figure \ref{fig: ch6_dhat khat} presents the estimates of the model parameter pair $(\hat{d}, \hat{K})$ for the 114 connectomes. The red dots represent the results from simultaneous model selection, and other colors are the results from BIC $\circ$ ZG. Consequently, there are four points for each graph, representing the pair of estimates by SMS, BIC $\circ$ ZG1, BIC $\circ$ ZG2 and BIC $\circ$ ZG3 respectively. 

For each graph and each specific algorithm, we have three ARIs indicating the clustering accuracy for three different cases. We are interested in how well our SMS algorithms perform compared with the traditional BIC $\circ$ ZG algorithms. As an example, Figure \ref{fig: ch6_hist difference of ARIs} shows the result of the paired difference of ARIs between SMS and the BIC $\circ$ ZG methods. Attribute hemisphere (left or right) is considered when computing ARI. Fixing two algorithms in competition, the differences of ARI are taken pairwise for all 114 graphs. We plot the histogram of those differences. More positive values in the histogram indicate stronger evidence that SMS outperforms BIC $\circ$ ZG, since higher ARI indicates that the clustering result is closer to the ground truth. From Figure \ref{fig: ch6_hist difference of ARIs}(a)-(c) we claim that the full SMS algorithm, dominates all BIC $\circ$ ZGs, following the observation that obviously more difference values are positive. In Figure \ref{fig: ch6_hist difference of ARIs}(d)-(f), although the number of positive values is close to that of negative ones, the reduced SMS algorithm still wins against the BIC $\circ$ ZGs slightly because of higher ARIs on average. Table \ref{tab: ch6_compare MCG/MCEG and ZGs} gives the results on all three attributes, where the number of graphs (out of 114) on which ARI of SMS is strictly larger than the sequential methods is reported in the column ``\#win''. Here we also consider the Louvain, Walktrap, and IRM methods in the competition. We again calculate p-values via the sign test.
The results show that the full simultaneous algorithm dominates in all cases against the BIC $\circ$ ZGs. 
Notice that the Louvain method demonstrates good performance for the hemisphere attribute, but it does not work well (with very small ARI values) for the tissue attribute. An analogous ``Two Truths'' phenomenon has been discovered in the work of \cite{priebe2018two}, where the authors find that Laplacian spectral embedding (LSE) better captures the hemisphere affinity structure while ASE better captures the tissue core-periphery structure. Here we see that Louvain, like LSE, is good for detecting the hemisphere affinity structure but not for detecting the tissue core-periphery structure. 
On the other hand, IRM performs well for the tissue attribute but poorly otherwise.

\begin{table}[hp]
    \centering
    \begin{tabular}{c|c|cc|cc}
        \toprule
         \multicolumn{2}{c|}{\multirow{2}{*}{}} & \multicolumn{2}{c|}{SMS} & \multicolumn{2}{c}{SMS-Reduced}   \\
         \hhline{~~----}
         \multicolumn{2}{c|}{} & \#win & p-value & \#win & p-value \\
         \hline
         \multirow{3}{*}{Hemisphere} & BIC $\circ$ ZG1 & 93 & 5.8e-13 & 66 & 0.037 \\
         & BIC $\circ$ ZG2 & 86 & 7.6e-9 & 62 & 0.151 \\
         & BIC $\circ$ ZG3 & 85 & 2.4e-8 & 64 & 0.080 \\
         & Louvain & 20 & 1 & 22 & 1 \\
         & Walktrap & 70 & 5.6e-3 & 52 & 0.800\\ 
         & IRM & 108 & 0 & 71 & 0.003\\ 
         \hline
         \multirow{3}{*}{Tissue} & BIC $\circ$ ZG1 & 69 & 0.009 & 41 & 0.998 \\
         & BIC $\circ$ ZG2 & 102 & 0 & 81 & 1.6e-6 \\
         & BIC $\circ$ ZG3 & 82 & 5.9e-7 & 56 & 0.537 \\
         & Louvain & 110 & 0 & 101 & 0 \\
         & Walktrap & 114 & 0 & 112 & 0 \\
         & IRM & 27 & 1 & 19 & 1\\ 
         \hline
         \multirow{3}{*}{4-block} & BIC $\circ$ ZG1 & 82 & 5.9e-7 & 57 & 0.463 \\
         & BIC $\circ$ ZG2 & 89 & 1.8e-10 & 60 & 0.256 \\
         & BIC $\circ$ ZG3 & 86 & 7.6e-9 & 67 & 0.024 \\
         & Louvain & 32 & 1 & 34 & 1 \\
         & Walktrap & 94 & 1.2e-13 & 59 & 0.320 \\
         & IRM & 102 & 0 & 92 & 2.7e-12\\ 
\bottomrule
    \end{tabular}
    \caption[The evidence that MCG/MCEG outperforms BIC $\circ$ ZG in terms of ARI]{The evidence that simultaneous model selection (SMS) outperforms the sequential BIC $\circ$ ZG in terms of adjusted Rand index (ARI), which is evaluated by three different versions of ground truth: hemisphere, tissue and the combination of the two (4-block). The number of graphs (out of 114) on which ARI of SMS is strictly larger than that of existing methods is reported in the column ``\#win''. The p-value for a sign test: $H_0: \theta \le 0.5$, $H_1: \theta > 0.5$, where $\theta$ is the probability that SMS wins, is reported next to the corresponding number. 
    }
    \label{tab: ch6_compare MCG/MCEG and ZGs}
\end{table}

\subsection{Computational considerations}

Naively, the eigenvector computation seems to have a cubic cost in the number of vertices.
In fact, spectral graph inference never needs the full spectrum; rather, only a few -- constant, or at most $o(n)$ -- eigenpairs are required.
(This is both a design consideration -- low-rank assumptions tend to be productive --
and a practical necessity, since an ill-conditioned adjacency matrix precludes getting all $n$ eigenpairs but does {\it not} preclude getting $o(n)$ eigenpairs.)
For a dense matrix with an eigengap, the cost for getting $O(1)$ eigenpairs is $O(n^2$). 

The real computational bottleneck for our simultaneous $(d,K)$ approach is the
multiple runs of the MLE algorithm (e.g., expectation-maximization, or EM).
When we run SMS,
the (truncated) eigenvector computation 
is done just once.
This pales in comparison to computational effort for the multiple EMs.
The Zhu-Ghodsi approach, while practically dominated by our method in terms of statistical performance, uses the same spectrum but then just $K_{max}$ EMs for ZG's single $\hat{d}$; SMS runs $K_{max}$ EMs for many possible dimensions.

The spectral computational aspects are well understood and, as describe above, a second-order issue $\dots$ practically.
Technically, the totality of the EM computations is bounded above by $O(n K_{max}^2 D^4)$, and in practice
for each $d$ we need run $K$ only up to the point at which the BIC decreases,
but this part -- and in fact the high-dimensional cases in this part -- still strongly dominates the spectral.

In summary:
for fixed $D$ and $K_{max}$ the spectral dominates theoretically even at $O(n^2)$, but this is a case where the theoretical ``Big-O'' analysis is misleading with respect to  practice: the MLEs dominate our computational complexity in practice.
However, our approach is the only computationally feasible simultaneous option. (The MCMC approach of \cite{heard2019} is, at present, not scalable.)

\section{Conclusion}
\label{sec: Conclusion}
This paper presents a novel simultaneous model selection framework -- dimensionality and cluster complexity -- specifically for vertex clustering on stochastic block model graphs.

In the first part of the paper, we propose the extended adjacency spectral embedding (extended ASE), in which the embedding is performed with a fixed (large) dimension. Under the framework of model-based clustering, we propose a family of Gaussian mixture models (GMM) to parameterize the entire extended ASE. The basis of the model is a state-of-the-art distributional result for the informative dimensions, as well as a conjecture founded on evidence from principled simulations for the redundant dimensions.

In the second part of the paper, we propose a simultaneous model selection (SMS) framework to improve upon sequential model selection. The framework is specifically tailored for the vertex clustering task on stochastic block model graphs. In contrast with sequential model selection, our approach identifies the embedding dimension and mixture complexity 
simultaneously. Moreover, we state and prove a theorem on the consistency of model parameter estimates. The theorem claims that the estimates in the model selection procedure given by our SMS method converge to the underlying truth for large graphs, provided the extended ASE follows the distribution in our proposed model. Based on SMS, we also develop two heuristic algorithms to solve the vertex clustering problem. The effectiveness of the algorithms is demonstrated in simulations and a real data experiment.

Note that the rank of $B$, the ``best'' embedding dimension for clustering, and the number of clusters interact in a rather complex manner. In Figure \ref{fig: ch3_inform},
the right plot shows that $d=2$ is likely to be the correct embedding dimension (and this will
be even more obvious for larger values of $n$) and yet both the rank of the $B$ matrix and
the number of blocks is $3$. Directly utilizing a measure of clustering performance in the
parameter selection and modeling is an area for further research.

We have focused on the so called ``hard clustering'' procedure in which each vertex is assigned
to a unique cluster. However, the use of Gaussian mixture model clustering allows for
``soft clustering'' whereby the likelihood ratio is used as the
assignment probability, rather taking the $\arg\max$ to provide a hard threshold.


\section*{Acknowledgements}
The authors thank two referees and an associate editor for their time and effort; this manuscript is significantly improved based on their thorough and constructive reports.
This work was partially supported by DARPA D3M contract FA8750-17-2-0112 and by the 
Naval Engineering Education Consortium through Office of Naval Research Award Number N00174-19-1-0011.

The required {\tt R} source codes and the data to reproduce the results in this manuscript can be found in \url{https://github.com/youngser/dhatkhat}.

\section{Appendix}
For the proof of Theorem \ref{thm: main theorem}, we begin with the following lemma.

\begin{lemma}
    \noindent
    Following the notation in Theorem \ref{thm: main theorem}, for all $d, K$,
    \begin{equation}
        \frac{1}{n} \sum_{i=1}^{n} \log\left[\frac{f(\hat{Z}^{(n)}_i; \theta^*(d_0, K_0))}{f(\hat{Z}^{(n)}_i; \hat\theta(d, K))}\right] \overset{p} \longrightarrow D_{\text{KL}}[f(\cdot ; \theta^*(d_0, K_0)) || f(\cdot ; \theta^*(d, K))]
    \end{equation}
    as $n \to \infty$.
    \label{lem: the lemma to proof main theorem}
\end{lemma}
\noindent \textbf{Proof}. By the definition of Kullback-Leibler divergence in (\ref{eq: KL divergence}),
\begin{align}
    &D_{\text{KL}}\left[f(\cdot ; \theta^*(d_0, K_0)) || f(\cdot ; \theta^*(d, K))\right] \notag\\
    = &\mathbb{E}\left[ \log\left( \frac{f(\hat{Z}^{(n)}_i; \theta^*(d_0, K_0))}{f(\hat{Z}^{(n)}_i; \theta^*(d, K))}\right) \right] \notag\\
    = &\mathbb{E}\left[\log(f(\hat{Z}^{(n)}_i; \theta^*(d_0, K_0)))\right] - \mathbb{E}\left[\log(f(\hat{Z}^{(n)}_i; \theta^*(d, K)))\right].
\end{align}
So we can prove the lemma by showing
\begin{equation}
  \label{lemma detail 1}
  \frac{1}{n} \sum_{i=1}^{n} \log\left[f(\hat{Z}^{(n)}_i; \theta^*(d_0, K_0))\right] \overset{p} \longrightarrow \mathbb{E}[\log(f(\hat{X}_i; \theta^*(d_0, K_0)))]
\end{equation}
and
\begin{equation}
  \label{lemma detail 2}
  \frac{1}{n} \sum_{i=1}^{n} \log\left[f(\hat{Z}^{(n)}_i; \hat\theta(d, K))\right] \overset{p} \longrightarrow \mathbb{E}[\log(f(\hat{X}_i; \theta^*(d, K)))]
\end{equation}
as $n \rightarrow \infty$. (\ref{lemma detail 1}) is the direct result of the law of large numbers. (\ref{lemma detail 2}) is the result of theorem 2.2 in \cite{white1982maximum} then followed by Slutsky's theorem. $\hfill\blacksquare$

Now we show the proof of Theorem \ref{thm: main theorem} as follows.

\noindent \textbf{Proof of Theorem \ref{thm: main theorem}}. Since $\hat{d}^{(n)}$ and $\hat{K}^{(n)}$ are both integer random variables, showing \\
$(\hat{d}^{(n)}, \hat{K}^{(n)}) \overset{p}{\longrightarrow} (d_0, K_0)$ is equivalent to showing
\begin{equation}
    \mathbb{P}\left[(\hat{d}^{(n)}, \hat{K}^{(n)}) = (d_0, K_0)\right] \longrightarrow 1.
    \label{eq: to show 1}
\end{equation}
By the definition of $\hat{d}^{(n)}$ and $\hat{K}^{(n)}$ in (\ref{eq: definition of dhat khat}), the event $\left\{(\hat{d}^{(n)}, \hat{K}^{(n)}) = (d_0, K_0)\right\}$ is equivalent to the event $\left\{(d_0, K_0) = \arg\max_{d \in [D], K \in [K_{\text{max}}]} \text{BIC} (\hat{Z}^{(n)}; d, K)\right\}$, which is equivalent to $\bigcap_{d,K} \left\{\text{BIC}(\hat{Z}^{(n)}; d_0, K_0) \ge \text{BIC} (\hat{Z}^{(n)}; d, K) \right\}$, so
\begin{align}
    &\mathbb{P}\left[(\hat{d}^{(n)}, \hat{K}^{(n)}) = (d_0, K_0)\right] \notag\\
    = &\mathbb{P}\left[ \bigcap_{d \in [D], K \in [K_{\text{max}}]} \left\{\text{BIC} (\hat{Z}^{(n)}; d_0, K_0) \ge \text{BIC} (\hat{Z}^{(n)}; d, K)\right\} \right] \notag\\
    = &1 - \mathbb{P}\left[ \bigcup_{d \in [D], K \in [K_{\text{max}}]} \left\{\text{BIC} (\hat{Z}^{(n)}; d_0, K_0) < \text{BIC} (\hat{Z}^{(n)}; d, K)\right\} \right]\notag\\
    \ge&1-\sum_{d \in [D], K \in [K_{\text{max}}]} \mathbb{P}\left[ \text{BIC} (\hat{Z}^{(n)}; d_0, K_0) < \text{BIC} (\hat{Z}^{(n)}; d, K) \right].
\end{align}
Thus in order to show (\ref{eq: to show 1}), it is sufficient to show
\begin{equation}
    \mathbb{P}\left[ \text{BIC} (\hat{Z}^{(n)}; d_0, K_0) < \text{BIC} (\hat{Z}^{(n)}; d, K) \right]\longrightarrow 0
    \label{eq: proof P bic goes 0}
\end{equation}
as $n \to \infty$ for all $(d, K) \ne (d_0, K_0)$.
By the notation in (\ref{eq: bic in theorem}) and (\ref{eq: MLE in theorem}), we notice
\begin{align*}
  &\frac{1}{2n} \left[\text{BIC} (\hat{Z}^{(n)}; d_0, K_0) - \text{BIC} (\hat{Z}^{(n)}; d, K)\right] \\
  = &\frac{1}{2n} \left(2 \sum_{i=1}^{n} \log\left[f(\hat{Z}^{(n)}_i; \hat\theta(d_0, K_0))\right] - \eta(d_0, K_0) \log(n) \right)\\
  &- \frac{1}{2n} \left(2\sum_{i=1}^{n} \log[f(\hat{Z}^{(n)}_i; \hat\theta(d, K))] - \eta(d, K) \log(n) \right)
\end{align*}
\begin{align}
    = &\frac{1}{n} \sum_{i=1}^{n} \log\left[\frac{f(\hat{Z}^{(n)}_i; \hat\theta(d_0, K_0))}{f(\hat{Z}^{(n)}_i; \theta^*(d_0, K_0))}\right] - \frac{1}{n} \sum_{i=1}^{n} \log\left[\frac{f(\hat{Z}^{(n)}_i; \hat\theta(d, K))}{f(\hat{Z}^{(n)}_i; \theta^*(d_0, K_0))}\right] \notag\\
    &+ \frac{1}{2n} [\eta(d, K) - \eta(d_0, K_0)] \log(n) \notag\\
    = &-\frac{1}{n} \sum_{i=1}^{n} \log\left[\frac{f(\hat{Z}^{(n)}_i; \theta^*(d_0, K_0))}{f(\hat{Z}^{(n)}_i; \hat\theta(d_0, K_0))}\right] \notag\\
    &+ \left( \frac{1}{n} \sum_{i=1}^{n} \log\left[\frac{f(\hat{Z}^{(n)}_i; \theta^*(d_0, K_0))}{f(\hat{Z}^{(n)}_i; \hat\theta(d, K))}\right] - D_{\text{KL}}[f(\cdot ; \theta^*(d_0, K_0)) || f(\cdot ; \theta^*(d, K))] \right) \notag\\
    &+ \left( D_{\text{KL}}[f(\cdot ; \theta^*(d_0, K_0)) || f(\cdot ; \theta^*(d, K))] + \frac{1}{2n} [\eta(d, K) - \eta(d_0, K_0)] \log(n) \right) \notag\\
    = &S_1 + S_2 + S_3
  \label{eq: proof s1 s2 s3}
\end{align}
where we let
\begin{align}
    \label{eq: s1}
    S_1 &= -\frac{1}{n} \sum_{i=1}^{n} \log\left[\frac{f(\hat{Z}^{(n)}_i; \theta^*(d_0, K_0))}{f(\hat{Z}^{(n)}_i; \hat\theta(d_0, K_0))}\right], \\
    \label{eq: s2}
    S_2 &= \frac{1}{n} \sum_{i=1}^{n} \log\left[\frac{f(\hat{Z}^{(n)}_i; \theta^*(d_0, K_0))}{f(\hat{Z}^{(n)}_i; \hat\theta(d, K))}\right] - D_{\text{KL}}[f(\cdot ; \theta^*(d_0, K_0)) || f(\cdot ; \theta^*(d, K))], \\
    \label{eq: s3}
    S_3 &= D_{\text{KL}}[f(\cdot ; \theta^*(d_0, K_0)) || f(\cdot ; \theta^*(d, K))] + \frac{1}{2n} [\eta(d, K) - \eta(d_0, K_0)] \log(n).
\end{align}
So for any $\epsilon > 0$, by (\ref{eq: proof s1 s2 s3}),
\begin{align}
  &\mathbb{P}\left[ \text{BIC} (\hat{Z}^{(n)}; d_0, K_0) < \text{BIC} (\hat{Z}^{(n)}; d, K) \right] \notag\\
  = &\mathbb{P}\left[\frac{1}{2n} \left[\text{BIC} (\hat{Z}^{(n)}; d_0, K_0) - \text{BIC} (\hat{Z}^{(n)}; d, K)\right] < 0 \right] \notag\\
  = &\mathbb{P}\left[S_1 + S_2 + S_3 < 0 \right] \notag\\
  \le &\mathbb{P}[S_1 < -\epsilon] + \mathbb{P}[S_2 < -\epsilon] + \mathbb{P}[S_3 < 2\epsilon].
  \label{eq: proof bic le P P P}
\end{align}
Here we use the fact that 
\begin{equation}
    \{S_1 + S_2 + S_3 < 0 \} \subset \left\{\{S_1 < -\epsilon\} \bigcup \{S_2 < -\epsilon\} \bigcup \{ S_3 < 2\epsilon\} \right\}.
\end{equation}
Thus
\begin{equation}
    \mathbb{P}\left[S_1 + S_2 + S_3 < 0 \right] \le \mathbb{P}[S_1 < -\epsilon] + \mathbb{P}[S_2 < -\epsilon] + \mathbb{P}[S_3 < 2\epsilon].
\end{equation}
Now in order to show (\ref{eq: proof P bic goes 0}), by (\ref{eq: proof bic le P P P}) it suffices to show
\begin{align}
    &\mathbb{P}[S_1 < -\epsilon] \longrightarrow 0,\\
    &\mathbb{P}[S_2 < -\epsilon] \longrightarrow 0,\\ 
    &\mathbb{P}[S_3 < 2\epsilon] \longrightarrow 0.
\end{align}
For (\ref{eq: s1}), by Lemma \ref{lem: the lemma to proof main theorem},
\begin{equation}
  \frac{1}{n} \sum_{i=1}^{n} \log\left[\frac{f(\hat{Z}^{(n)}_i; \theta^*(d_0, K_0))}{f(\hat{Z}^{(n)}_i; \hat\theta(d_0, K_0))}\right] \overset{p} \longrightarrow D_{\text{KL}}[f(\cdot ; \theta^*(d_0, K_0)) || f(\cdot ; \theta^*(d_0, K_0))] = 0
\end{equation}
so
\begin{equation}
    \mathbb{P}\left[ -\frac{1}{n} \sum_{i=1}^{n} \log\left[\frac{f(\hat{Z}^{(n)}_i; \theta^*(d_0, K_0))}{f(\hat{Z}^{(n)}_i; \hat\theta(d_0, K_0))}\right] < -\epsilon\right] \longrightarrow 0.
    \label{eq: proof s1}
\end{equation}
For (\ref{eq: s2}), also by Lemma \ref{lem: the lemma to proof main theorem},
\begin{equation}
    \frac{1}{n} \sum_{i=1}^{n} \log\left[\frac{f(\hat{Z}^{(n)}_i; \theta^*(d_0, K_0))}{f(\hat{Z}^{(n)}_i; \hat\theta(d, K))}\right] \overset{p} \longrightarrow D_{\text{KL}}[f(\cdot ; \theta^*(d_0, K_0)) || f(\cdot ; \theta^*(d, K))]
\end{equation}
so
\begin{equation}
    \mathbb{P}\left[ \frac{1}{n} \sum_{i=1}^{n} \log\left[\frac{f(\hat{Z}^{(n)}_i; \theta^*(d_0, K_0))}{f(\hat{Z}^{(n)}_i; \hat\theta(d, K))}\right] - D_{\text{KL}}[f(\cdot ; \theta^*(d_0, K_0)) || f(\cdot ; \theta^*(d, K))] < -\epsilon \right] \longrightarrow 0.
    \label{eq: proof s2}
\end{equation}
For (\ref{eq: s3}), if $(d, K) \ne (d_0, K_0)$, then by the identifiability assumption (b), we know
\begin{equation}
    D_{\text{KL}}[f(\cdot ; \theta^*(d_0, K_0)) || f(\cdot ; \theta^*(d, K))] > 0.
\end{equation}
Thus if we take $\epsilon = \frac{1}{3} D_{\text{KL}}[f(\cdot ; \theta^*(d_0, K_0)) || f(\cdot ; \theta^*(d, K))]$, then we have
\begin{equation}
    \mathbb{P}\left[ D_{\text{KL}}[f(\cdot ; \theta^*(d_0, K_0)) || f(\cdot ; \theta^*(d, K))] + \frac{1}{2n} [\eta(d, K) - \eta(d_0, K_0)] \log(n) < 2 \epsilon \right] \longrightarrow 0
    \label{eq: proof s3}
\end{equation}
because $\frac{\log(n)}{2n} \longrightarrow 0$ as $n \longrightarrow \infty$. Combining (\ref{eq: proof s1}), (\ref{eq: proof s2}) and (\ref{eq: proof s3}), we have
\begin{equation}
    \mathbb{P}\left[\frac{1}{2n} [\text{BIC} (\hat{Z}^{(n)}; d_0, K_0) - \text{BIC} (\hat{Z}^{(n)}; d, K)] < 0\right] \longrightarrow 0
\end{equation}
as $n \longrightarrow \infty$ for all $d \in [D]$ and $K \in K_{\text{max}}$. So we have shown (\ref{eq: proof P bic goes 0}), which finishes the proof of
\begin{equation}
  (\hat{d}^{(n)}, \hat{K}^{(n)}) \overset{p}{\longrightarrow} (d_0, K_0) 
\end{equation}
as $n \longrightarrow \infty$. $\hfill\blacksquare$

\bibliography{reference.bib}

\begin{thebibliography}{}

\bibitem[Akaike, 1998]{akaike1998information}
Akaike, H. (1998).
\newblock Information theory and an extension of the maximum likelihood
  principle.
\newblock In {\em Selected Papers of Hirotugu Akaike}, pages 199--213.
  Springer.

\bibitem[Athreya et~al., 2018]{athreya2017statistical}
Athreya, A., Fishkind, D.~E., Tang, M., Priebe, C.~E., Park, Y., Vogelstein,
  J.~T., Levin, K., Lyzinski, V., Qin, Y., and Sussman, D.~L. (2018).
\newblock Statistical inference on random dot product graphs: a survey.
\newblock {\em Journal of Machine Learning Research}, 18(226):1--92.

\bibitem[Athreya et~al., 2016]{athreya2016limit}
Athreya, A., Priebe, C.~E., Tang, M., Lyzinski, V., Marchette, D.~J., and
  Sussman, D.~L. (2016).
\newblock A limit theorem for scaled eigenvectors of random dot product graphs.
\newblock {\em Sankhya A}, 78(1):1--18.

\bibitem[Bickel and Chen, 2009]{bickel2009nonparametric}
Bickel, P.~J. and Chen, A. (2009).
\newblock A nonparametric view of network models and newman--girvan and other
  modularities.
\newblock {\em Proceedings of the National Academy of Sciences},
  106(50):21068--21073.

\bibitem[Biernacki et~al., 2000]{biernacki2000assessing}
Biernacki, C., Celeux, G., and Govaert, G. (2000).
\newblock Assessing a mixture model for clustering with the integrated
  completed likelihood.
\newblock {\em IEEE Transactions on Pattern Analysis and Machine Intelligence},
  22(7):719--725.

\bibitem[Blondel et~al., 2008]{blondel2008fast}
Blondel, V.~D., Guillaume, J.-L., Lambiotte, R., and Lefebvre, E. (2008).
\newblock Fast unfolding of communities in large networks.
\newblock {\em Journal of Statistical Mechanics: Theory and Experiment},
  2008(10):P10008.

\bibitem[Bullmore and Bassett, 2011]{bullmore2011brain}
Bullmore, E.~T. and Bassett, D.~S. (2011).
\newblock Brain graphs: graphical models of the human brain connectome.
\newblock {\em Annual Review of Clinical Psychology}, 7:113--140.

\bibitem[Campbell et~al., 1999]{campbell1999model}
Campbell, J.~G., Fraley, C., Stanford, D., Murtagh, F., and Raftery, A.~E.
  (1999).
\newblock Model-based methods for textile fault detection.
\newblock {\em International Journal of Imaging Systems and Technology},
  10(4):339--346.

\bibitem[Cape et~al., 2019]{CAPENETSCI}
Cape, J., Tang, M., and Priebe, C.~E. (2019).
\newblock {On spectral embedding performance and elucidating network structure
  in stochastic blockmodel graphs}.
\newblock {\em Network Science}, 7(3):269--291.

\bibitem[Celeux and Soromenho, 1996]{celeux1996entropy}
Celeux, G. and Soromenho, G. (1996).
\newblock An entropy criterion for assessing the number of clusters in a
  mixture model.
\newblock {\em Journal of Classification}, 13(2):195--212.

\bibitem[Danon et~al., 2005]{danon2005comparing}
Danon, L., Diaz-Guilera, A., Duch, J., and Arenas, A. (2005).
\newblock Comparing community structure identification.
\newblock {\em Journal of Statistical Mechanics: Theory and Experiment},
  2005(09):P09008.

\bibitem[Dasgupta and Raftery, 1998]{dasgupta1998detecting}
Dasgupta, A. and Raftery, A.~E. (1998).
\newblock Detecting features in spatial point processes with clutter via
  model-based clustering.
\newblock {\em Journal of the American Statistical Association},
  93(441):294--302.

\bibitem[Erd\H{o}s and R{\'e}nyi, 1960]{erdos1960evolution}
Erd\H{o}s, P. and R{\'e}nyi, A. (1960).
\newblock On the evolution of random graphs.
\newblock {\em Publ. Math. Inst. Hung. Acad. Sci}, 5(1):17--61.

\bibitem[Fop and Murphy, 2018]{fop2018variable}
Fop, M. and Murphy, T.~B. (2018).
\newblock Variable selection methods for model-based clustering.
\newblock {\em Statistics Surveys}, 12:18--65.

\bibitem[Fraley and Raftery, 2002]{fraley2002model}
Fraley, C. and Raftery, A.~E. (2002).
\newblock Model-based clustering, discriminant analysis, and density
  estimation.
\newblock {\em Journal of the American statistical Association},
  97(458):611--631.

\bibitem[Girvan and Newman, 2002]{girvan2002community}
Girvan, M. and Newman, M.~E. (2002).
\newblock Community structure in social and biological networks.
\newblock {\em Proceedings of the National Academy of Sciences},
  99(12):7821--7826.

\bibitem[Handcock et~al., 2007]{Handcock2007ModelbasedCF}
Handcock, M., Raftery, A., and Tantrum, J. (2007).
\newblock Model-based clustering for social networks.
\newblock {\em Journal of the Royal Statistical Society: Series A (Statistics
  in Society)}, 170:301 -- 354.

\bibitem[Hardy, 1996]{hardy1996number}
Hardy, A. (1996).
\newblock On the number of clusters.
\newblock {\em Computational Statistics \& Data Analysis}, 23(1):83--96.

\bibitem[Hoff, 2005]{hoff2005}
Hoff, P.~D. (2005).
\newblock Bilinear mixed-effects models for dyadic data.
\newblock {\em Journal of the American Statistical Association},
  100(469):286--295.

\bibitem[Hoff, 2008]{hoff2008}
Hoff, P.~D. (2008).
\newblock Modeling homophily and stochastic equivalence in symmetric relational
  data.
\newblock {\em Advances in Neural Information Processing Systems},
  (20):657--664.

\bibitem[Hoff et~al., 2002]{hoff2002latent}
Hoff, P.~D., Raftery, A.~E., and Handcock, M.~S. (2002).
\newblock Latent space approaches to social network analysis.
\newblock {\em Journal of the American Statistical Association},
  97(460):1090--1098.

\bibitem[Holland et~al., 1983]{holland1983stochastic}
Holland, P.~W., Laskey, K.~B., and Leinhardt, S. (1983).
\newblock Stochastic blockmodels: First steps.
\newblock {\em Social Networks}, 5(2):109--137.

\bibitem[Hubert and Arabie, 1985]{hubert1985comparing}
Hubert, L. and Arabie, P. (1985).
\newblock Comparing partitions.
\newblock {\em Journal of Classification}, 2(1):193--218.

\bibitem[Jaccard, 1912]{jaccard1912distribution}
Jaccard, P. (1912).
\newblock The distribution of the flora in the alpine zone. 1.
\newblock {\em New Phytologist}, 11(2):37--50.

\bibitem[Jackson, 1993]{jackson1993stopping}
Jackson, D.~A. (1993).
\newblock Stopping rules in principal components analysis: a comparison of
  heuristical and statistical approaches.
\newblock {\em Ecology}, 74(8):2204--2214.

\bibitem[Jain et~al., 2000]{jain2000statistical}
Jain, A.~K., Duin, R.~P., and Mao, J. (2000).
\newblock Statistical pattern recognition: A review.
\newblock {\em IEEE Transactions on Pattern Analysis and Machine Intelligence},
  22(1):4--37.

\bibitem[James et~al., 2001]{JCP2001}
James, L.~F., Priebe, C.~E., and Marchette, D.~J. (2001).
\newblock Consistent estimation of mixture complexity.
\newblock {\em The Annals of Statistics}, 29(5):1281--1296.

\bibitem[Jolliffe, 2011]{jolliffe2011principal}
Jolliffe, I. (2011).
\newblock Principal component analysis.
\newblock In {\em International Encyclopedia of Statistical Science}, pages
  1094--1096. Springer.

\bibitem[Kemp et~al., 2006]{irm}
Kemp, C., Tenenbaum, J.~B., Griffiths, T.~L., Yamada, T., and Ueda, N. (2006).
\newblock Learning systems of concepts with an infinite relational model.
\newblock In {\em Proceedings of the 21st National Conference on Artificial
  Intelligence - Volume 1}, AAAI’06, pages 381--388. AAAI Press.

\bibitem[Keribin, 2000]{keribin2000consistent}
Keribin, C. (2000).
\newblock Consistent estimation of the order of mixture models.
\newblock {\em Sankhy{\=a}: The Indian Journal of Statistics, Series A},
  62(1):49--66.

\bibitem[Kiar et~al., 2018]{kiar2018high}
Kiar, G., Bridgeford, E.~W., Gray~Roncal, W.~R., for Reliability, C., (CoRR),
  R., Chandrashekhar, V., Mhembere, D., Ryman, S., Zuo, X.-N., Margulies,
  D.~S., Craddock, R.~C., Priebe, C.~E., Jung, R., Calhoun, V.~D., Caffo, B.,
  Burns, R., Milham, M.~P., and Vogelstein, J.~T. (2018).
\newblock A high-throughput pipeline identifies robust connectomes but
  troublesome variability.
\newblock {\em bioRxiv preprint, doi:10.1101/188706}.

\bibitem[Lazer et~al., 2009]{lazer2009life}
Lazer, D., Pentland, A., Adamic, L., Aral, S., Barabasi, A.-L., Brewer, D.,
  Christakis, N., Contractor, N., Fowler, J., Gutmann, M., Jebara, T., King,
  G., Macy, M., Roy, D., and Van~Alstyne, M. (2009).
\newblock {Life in the network: the coming age of computational social
  science}.
\newblock {\em Science}, 323(5915):721--723.

\bibitem[Lei et~al., 2015]{lei2015consistency}
Lei, J., Rinaldo, A., et~al. (2015).
\newblock Consistency of spectral clustering in stochastic block models.
\newblock {\em The Annals of Statistics}, 43(1):215--237.

\bibitem[Lyzinski et~al., 2017]{lyzinski2017community}
Lyzinski, V., Tang, M., Athreya, A., Park, Y., and Priebe, C.~E. (2017).
\newblock Community detection and classification in hierarchical stochastic
  blockmodels.
\newblock {\em IEEE Transactions on Network Science and Engineering},
  4(1):13--26.

\bibitem[Meil{\u{a}}, 2007]{meilua2007comparing}
Meil{\u{a}}, M. (2007).
\newblock Comparing clusterings---an information based distance.
\newblock {\em Journal of Multivariate Analysis}, 98(5):873--895.

\bibitem[Milligan and Cooper, 1985]{milligan1985examination}
Milligan, G.~W. and Cooper, M.~C. (1985).
\newblock An examination of procedures for determining the number of clusters
  in a data set.
\newblock {\em Psychometrika}, 50(2):159--179.

\bibitem[Newman, 2006]{newman2006modularity}
Newman, M.~E. (2006).
\newblock Modularity and community structure in networks.
\newblock {\em Proceedings of the National Academy of Sciences},
  103(23):8577--8582.

\bibitem[Newman and Girvan, 2004]{newman2004finding}
Newman, M.~E. and Girvan, M. (2004).
\newblock Finding and evaluating community structure in networks.
\newblock {\em Physical Review E}, 69(2):026113.

\bibitem[Pisano et~al., 2020]{ZPcurved}
Pisano, Z.~M., Agterberg, J.~S., Priebe, C.~E., and Naiman, D.~Q. (2020).
\newblock Spectral graph clustering via the expectation-solution algorithm.
\newblock {\em arXiv preprint arXiv:2003.13462}.

\bibitem[Pons and Latapy, 2005]{pons2005computing}
Pons, P. and Latapy, M. (2005).
\newblock Computing communities in large networks using random walks.
\newblock In {\em International Symposium on Computer and Information
  Sciences}, pages 284--293. Springer.

\bibitem[Priebe et~al., 2019]{priebe2018two}
Priebe, C.~E., Park, Y., Vogelstein, J.~T., Conroy, J.~M., Lyzinskic, V., Tang,
  M., Athreya, A., Cape, J., and Bridgeford, E. (2019).
\newblock On a two truths phenomenon in spectral graph clustering.
\newblock {\em Proceedings of the National Academy of Sciences},
  116(13):5995--6000.

\bibitem[Proulx et~al., 2005]{proulx2005network}
Proulx, S.~R., Promislow, D.~E., and Phillips, P.~C. (2005).
\newblock Network thinking in ecology and evolution.
\newblock {\em Trends in Ecology \& Evolution}, 20(6):345--353.

\bibitem[Qin and Rohe, 2013]{qin2013regularized}
Qin, T. and Rohe, K. (2013).
\newblock Regularized spectral clustering under the degree-corrected stochastic
  blockmodel.
\newblock In {\em Advances in Neural Information Processing Systems}, pages
  3120--3128.

\bibitem[Raftery and Dean, 2006]{raftery2006variable}
Raftery, A.~E. and Dean, N. (2006).
\newblock Variable selection for model-based clustering.
\newblock {\em Journal of the American Statistical Association},
  101(473):168--178.

\bibitem[Roeder and Wasserman, 1997]{roeder1997practical}
Roeder, K. and Wasserman, L. (1997).
\newblock Practical {B}ayesian density estimation using mixtures of normals.
\newblock {\em Journal of the American Statistical Association},
  92(439):894--902.

\bibitem[Rohe et~al., 2011]{rohe2011spectral}
Rohe, K., Chatterjee, S., Yu, B., et~al. (2011).
\newblock Spectral clustering and the high-dimensional stochastic blockmodel.
\newblock {\em The Annals of Statistics}, 39(4):1878--1915.

\bibitem[Rosvall and Bergstrom, 2008]{rosvall2008maps}
Rosvall, M. and Bergstrom, C.~T. (2008).
\newblock Maps of random walks on complex networks reveal community structure.
\newblock {\em Proceedings of the National Academy of Sciences},
  105(4):1118--1123.

\bibitem[Rubin-Delanchy et~al., 2017]{rubin2017statistical}
Rubin-Delanchy, P., Priebe, C.~E., Tang, M., and Cape, J. (2017).
\newblock A statistical interpretation of spectral embedding: the generalised
  random dot product graph.
\newblock {\em arXiv preprint arXiv:1709.05506}.

\bibitem[{Sanna Passino} and {Heard}, 2020]{heard2019}
{Sanna Passino}, F. and {Heard}, N.~A. (2020).
\newblock {Bayesian estimation of the latent dimension and communities in
  stochastic blockmodels}.
\newblock {\em Statistics and Computing}, to appear.

\bibitem[Schwarz et~al., 1978]{schwarz1978estimating}
Schwarz, G. et~al. (1978).
\newblock Estimating the dimension of a model.
\newblock {\em The Annals of Statistics}, 6(2):461--464.

\bibitem[Scrucca et~al., 2016]{mclust}
Scrucca, L., Fop, M., Murphy, T.~B., and Raftery, A.~E. (2016).
\newblock {mclust} 5: clustering, classification and density estimation using
  {G}aussian finite mixture models.
\newblock {\em The {R} Journal}, 8(1):289--317.

\bibitem[Smyth, 2000]{smyth2000model}
Smyth, P. (2000).
\newblock Model selection for probabilistic clustering using cross-validated
  likelihood.
\newblock {\em Statistics and Computing}, 10(1):63--72.

\bibitem[Stanford and Raftery, 1997]{stanford1997principal}
Stanford, D. and Raftery, A.~E. (1997).
\newblock Principal curve clustering with noise.
\newblock Technical Report 317, University of Washington.

\bibitem[Sussman et~al., 2012]{sussman2012consistent}
Sussman, D.~L., Tang, M., Fishkind, D.~E., and Priebe, C.~E. (2012).
\newblock A consistent adjacency spectral embedding for stochastic blockmodel
  graphs.
\newblock {\em Journal of the American Statistical Association},
  107(499):1119--1128.

\bibitem[Sussman et~al., 2014]{sussman2014consistent}
Sussman, D.~L., Tang, M., and Priebe, C.~E. (2014).
\newblock Consistent latent position estimation and vertex classification for
  random dot product graphs.
\newblock {\em IEEE Transactions on Pattern Analysis and Machine Intelligence},
  36(1):48--57.

\bibitem[Tang et~al., 2017]{TangAsyEff}
Tang, M., Cape, J., and Priebe, C.~E. (2017).
\newblock Asymptotically efficient estimators for stochastic blockmodels: the
  naive {MLE}, the rank-constrained {MLE}, and the spectral.
\newblock {\em arXiv preprint arXiv:1710.10936}.

\bibitem[Tang et~al., 2018]{tang2018limit}
Tang, M., Priebe, C.~E., et~al. (2018).
\newblock Limit theorems for eigenvectors of the normalized laplacian for
  random graphs.
\newblock {\em The Annals of Statistics}, 46(5):2360--2415.

\bibitem[Von~Luxburg, 2007]{von2007tutorial}
Von~Luxburg, U. (2007).
\newblock A tutorial on spectral clustering.
\newblock {\em Statistics and Computing}, 17(4):395--416.

\bibitem[Ward et~al., 2011]{ward2011network}
Ward, M.~D., Stovel, K., and Sacks, A. (2011).
\newblock Network analysis and political science.
\newblock {\em Annual Review of Political Science}, 14:245--264.

\bibitem[White, 1982]{white1982maximum}
White, H. (1982).
\newblock Maximum likelihood estimation of misspecified models.
\newblock {\em Econometrica: Journal of the Econometric Society}, 50(1):1--25.

\bibitem[Young and Scheinerman, 2007]{young2007random}
Young, S.~J. and Scheinerman, E.~R. (2007).
\newblock Random dot product graph models for social networks.
\newblock In {\em International Workshop on Algorithms and Models for the
  Web-Graph}, pages 138--149. Springer.

\bibitem[Zhu and Ghodsi, 2006]{zhu2006automatic}
Zhu, M. and Ghodsi, A. (2006).
\newblock Automatic dimensionality selection from the scree plot via the use of
  profile likelihood.
\newblock {\em Computational Statistics \& Data Analysis}, 51(2):918--930.

\end{thebibliography}
\end{document}